\documentclass[11pt]{article}
\usepackage[utf8]{inputenc}
\usepackage[T1]{fontenc}
\usepackage{lmodern}
\usepackage[margin=1in]{geometry}
\usepackage{amsmath,amssymb}
\usepackage{graphicx}
\usepackage{longtable,booktabs,array}
\usepackage{calc}
\usepackage{etoolbox}
\usepackage{xcolor}
\usepackage{caption}
\usepackage{pdflscape}
\usepackage{newunicodechar}
\usepackage[hidelinks]{hyperref}
\providecommand{\real}[1]{#1}

\newunicodechar{κ}{\ensuremath{\kappa}}\newunicodechar{Δ}{\ensuremath{\Delta}}
\newunicodechar{γ}{\ensuremath{\gamma}}\newunicodechar{α}{\ensuremath{\alpha}}
\newunicodechar{β}{\ensuremath{\beta}}\newunicodechar{μ}{\ensuremath{\mu}}
\newunicodechar{≥}{\ensuremath{\geq}}\newunicodechar{≤}{\ensuremath{\leq}}
\newunicodechar{×}{\ensuremath{\times}}\newunicodechar{±}{\ensuremath{\pm}}
\newunicodechar{−}{\ensuremath{-}}\newunicodechar{·}{\ensuremath{\cdot}}
\newunicodechar{→}{\ensuremath{\rightarrow}}\newunicodechar{≈}{\ensuremath{\approx}}
\newunicodechar{∞}{\ensuremath{\infty}}\newunicodechar{γ}{\ensuremath{\gamma}}
\newunicodechar{–}{\textendash}\newunicodechar{—}{\textemdash}
\newunicodechar{‐}{-}\newunicodechar{‑}{-}\newunicodechar{†}{\textdagger}
\newunicodechar{’}{'}\newunicodechar{‘}{'}\newunicodechar{“}{``}\newunicodechar{”}{''}
\newunicodechar{…}{\ldots}
\begin{document}

\begin{center}
{\LARGE\bfseries Deep Learning Pose Estimation for Multi-Label Phenotyping of Co-Occurring Hyperkinetic Movement Disorders}\\[0.6em]
{\itshape A single-centre, exploratory proof-of-concept study using a pose-based hybrid pipeline combining deep learning pose estimation with feature-engineered supervised classification.}\\[1em]
{\large Laura Cif \textsuperscript{1,~2,} *, Diane Demailly \textsuperscript{2,~3}, Gabriella A. Horvàth \textsuperscript{4}, Juan Dario Ortigoza Escobar \textsuperscript{5,~6,~7}, Nathalie Dorison \textsuperscript{8}, Mayté Castro Jiménez \textsuperscript{1}, Cécile A. Hubsch \textsuperscript{1}, Thomas Wirth \textsuperscript{9,~10,~11}, Gun-Marie Hariz \textsuperscript{12}, Sophie Huby \textsuperscript{13}, Morgan Dornadic \textsuperscript{13}, Zohra Souei \textsuperscript{2,~14}, Muhammad Mushhood Ur Rehman \textsuperscript{15}, Simone Hemm \textsuperscript{16}, Mehdi Boulaymen \textsuperscript{2}, Eduardo M. Moraud \textsuperscript{17,~18}, Jocelyne Bloch \textsuperscript{18,~19}, Xavier Vasques \textsuperscript{2}}
\end{center}

{\footnotesize\begin{flushleft}
\textsuperscript{1}Service of Neurology, Department of Clinical Neurosciences, Lausanne University Hospital (CHUV) and University of Lausanne (UNIL), Lausanne, Switzerland\\
\textsuperscript{2}Institut du Neurone, Montferrier sur Lez, France\\
\textsuperscript{3}Department of Neurology, Clinique Beau Soleil, Institut Mutualiste Montpelliérain, Montpellier, France\\
\textsuperscript{4}Department of Pediatrics, British Columbia Children's Hospital, Vancouver, British Columbia, Canada\\
\textsuperscript{5}Movement Disorders Unit, Pediatric Neurology Department, Institut de Recerca, Hospital Sant Joan de Déu, Barcelona, Spain\\
\textsuperscript{6}European Reference Network for Rare Neurological Diseases (ERN-RND), Barcelona, Spain\\
\textsuperscript{7}U-703 Centre for Biomedical Research on Rare Diseases (CIBER-ER), Instituto de Salud Carlos III, Barcelona, Spain\\
\textsuperscript{8}Pediatric Neurosurgery~Department, CCMR Neurogenetique, European Reference Network~Brainteam Member, Rothschild Foundation Hospital, Paris,~France\\
\textsuperscript{9}Department of Neurology, University Hospital of Strasbourg, Strasbourg, France\\
\textsuperscript{10} Strasbourg Neuroscience Institute, Strasbourg University, Strasbourg, France\\
\textsuperscript{11} Institute of Genetics and Cellular and Molecular Biology, INSERM-U964, CNRS-UMR7104, University of Strasbourg, Illkirch-Graffenstaden, France\\
\textsuperscript{12}Department of Clinical Neuroscience, Umeå University, Umeå, Sweden\\
\textsuperscript{13}Department of Neurology, CHU Montpellier, Montpellier, France\\
\textsuperscript{14}Department of Neurosurgery, Fattouma Bourguiba University Hospital, Tunisia\\
\textsuperscript{15}University of Edinburgh, Scotland\\
\textsuperscript{16}Institute for Medical Engineering and Medical Informatics, School of Life Sciences, University of Applied Sciences and Arts Northwestern Switzerland, Muttenz, Switzerland\\
\textsuperscript{17}Department of Clinical Neurosciences, University Hospital Lausanne (CHUV), Lausanne, Switzerland\\
\textsuperscript{18}Defitech Center for Interventional Neurotherapies (NeuroRestore), University Hospital Lausanne and Ecole Polytechnique Fédérale de Lausanne, Lausanne, Switzerland\\
\textsuperscript{19}Department of Neurosurgery, Lausanne University Hospital (CHUV) and University of Lausanne (UNIL), Lausanne, Switzerland.\\[0.6em]
\noindent$^{*}$Correspondence: Laura Cif, \href{mailto:lauracif@institutduneurone.fr}{lauracif@institutduneurone.fr}
\end{flushleft}}

\begin{abstract}
\textbf{Objective:} To explore, in an single-centre proof-of-concept study, whether routine outpatient video combined with deep learning-based pose estimation and clinically interpretable kinematic features can support patient-level multi-label phenotyping of co-occurring hyperkinetic movement disorders.

\textbf{Methods:} In this exploratory single-centre proof-of-concept study, we analysed videos from 25 participants (21 patients with isolated or combined hyperkinetic movement disorders and 4 healthy controls) recorded during standardized outpatient video examinations. Videos were processed with deep learning--based markerless pose estimation (YOLOv8), and 2-dimensional keypoint trajectories were transformed into clinically interpretable kinematic descriptors spanning statistical, temporal, spectral, and complexity domains. Ten-second windows were aligned to consensus expert annotations for 8 hyperkinetic phenotypes: dystonia, tremor, myoclonus, chorea, athetosis, tics, ballismus, and stereotypies. Conventional supervised classifiers (gradient-boosted decision trees, support-vector machines, logistic regression, random forests, k-nearest neighbours, and a multilayer perceptron) were trained on these tabular descriptors. Window-level predictions were aggregated to the patient level, and label-specific decision thresholds were tuned on training participants only. Patient-level multi-label performance was reported under (i) the best single pipeline by macro-AUPRC, (ii) the best single pipeline by Hamming accuracy, (iii) prespecified nested cross-validation with per-label model selection within training folds (primary analysis), and (iv) post-hoc per-label best (exploratory upper bound).

\textbf{Results:} In patient-level multi-label evaluation, the best single pipeline selected by discrimination (StandardScaler + MLP) achieved a macro-average precision--recall area under the curve of 0.821 ± 0.019 and a macro--receiver operating characteristic area under the curve of 0.830 ± 0.029. The best single pipeline selected by Hamming accuracy (MinMaxScaler + SVM) reached 0.764 ± 0.041. Under prespecified nested cross-validation with per-label model selection (primary analysis), macro-AUPRC was 0.717 ± 0.030, macro-AUROC was 0.767 ± 0.069, Hamming accuracy was 0.764 ± 0.014, and patient--label agreement was 153/200 (76.5\%). Post-hoc per-label selection of the best-performing pipeline defined an exploratory upper bound of 172/200 (86.0\%).

\textbf{Interpretation:} In this small, single-centre, exploratory cohort, a pose-based hybrid pipeline combining deep learning pose estimation with feature-engineered supervised classification produced encouraging patient-level multi-label performance for co-occurring hyperkinetic movement disorders, with prespecified nested cross-validation yielding patient--label agreement of 153/200 (76.5\%) and an upper-bound estimate from per-label best models of 172/200 (86.0\%). These findings should be regarded as proof-of-concept; external, multicentre, prospective validation is required before any clinical or trial use can be considered.
\end{abstract}

\section{Introduction}

In both, adult and pediatric neurology, movement disorders (MDs) are broadly classified from a phenomenological perspective as either hyperkinetic or hypokinetic MDs. Accurate, updated epidemiological data on MDs help to raise awareness, improve recognition, and facilitate timely diagnosis and treatment that remains especially symptomatic \textsuperscript{1}. Despite the progress achieved for their clinical characterization and genetic diagnostic \textsuperscript{2,3,4}, the recognition of MDs phenomenology remains challenging during clinical examination with disagreement even between MDs specialists, particularly for the complex presentations of many neurodevelopmental, neurodegenerative and monogenic disorders. The current classification of genetic MDs distinguishes between isolated, combined, and mixed categories \textsuperscript{5}.

Traditionally, clinicians identify hyperkinetic MDs through direct observation and assess them by using standardized clinical rating scales that usually focus on individual motor phenomena (e.g., scales for dystonia, tremor or tics). Despite their widespread use, they may not capture the full spectrum of motor phenomenology for each patient \textsuperscript{6,7} and have inherent limitations, notably inter-rater variability, limited reproducibility, and the need for specialized training and expertise \textsuperscript{8}. Additionally, some scales focus on specific diseases \textsuperscript{9} without potential for generalization or application to other conditions. By contrast, other scales \textsuperscript{10} may have broader applicability to other developmental disorders. Currently, no available composite assessment tools enable the simultaneous identification and monitoring of MDs over different diseases associating multiple hyperkinetic MDs and through lifespan, in both children and adults. Accurate and objective assessment of hyperkinetic MDs is critical not only for diagnosis, but also for monitoring, facilitating effective patient care, and the development of targeted, symptom-specific interventions \textsuperscript{11}.

Recent advances in machine learning (ML) \textsuperscript{12} enable objective movement analysis from noninvasive clinical videos via markerless pose estimation. This provides anatomically grounded time series that can support scalable phenotyping beyond subjective bedside ratings \textsuperscript{8,13,~14} \textsuperscript{15}. Research leveraging deep learning (DL) has shown progress in various clinical applications, including gait analysis, rehabilitation assessment \textsuperscript{8}, tic detection and tremor quantification \textsuperscript{16}.

Despite various developments, existing ML approaches for detecting hyperkinetic MDs remain limited in scope, focusing on binary classification of individual hyperkinetic MDs and frequently restricted to short, task-specific video segments \textsuperscript{8}. While most existing ML models for MD analysis rely on shallow feature representations, primarily using raw joint coordinates or velocity, they often fail to capture the temporal complexity and variability inherent to a pathological, involuntary movement \textsuperscript{17--22}. One unmet need remains the identification and quantification of combined MDs, both in neurology outpatient clinics and through remote assessments, to support disease monitoring and therapeutic decision-making. The field prospect is to extend DL developments for enhanced accuracy of phenotype classification, the cornerstone of both the diagnostic and treatment process \textsuperscript{23}.

We therefore developed and evaluated, in a single-centre proof-of-concept cohort, a pose-based hybrid pipeline that couples deep learning--based markerless pose estimation with feature engineering of clinically interpretable kinematic descriptors and conventional supervised classifiers (gradient-boosted decision trees, support-vector machines, logistic regression, random forests, k-nearest neighbours, and a multilayer perceptron). We evaluated, in a hierarchy of analyses with increasing clinical complexity, (i) window-level binary screening of symptom-expressing patient windows against healthy control windows, (ii) patient-level multi-label phenotyping via p90 percentile aggregation of window probabilities with control-aware thresholding, and (iii) decision-level interpretability linking predictions to kinematic families and anatomical regions. Because deep learning is used only in the upstream pose-estimation step and the downstream classification operates on tabular kinematic summaries computed within each 10-s window, the framework is best characterised as feature engineering--based supervised multi-label classification rather than as true sequential or temporal deep learning.

\section{Materials And Methods}

\textbf{Participant Recruitment and Video Acquisition}

Twenty-one consecutive patients (range 17-75 years, 12 females) diagnosed with isolated or combined MDs including dystonia and associated hyperkinetic MDs (i.e.~tremor, myoclonus, chorea, tics, athetosis, stereotypies, ballismus) and four healthy controls were enrolled in the study. The demographics and clinical characteristics are detailed in \textbf{(Table 1)}. Video recordings were prospectively collected from routine clinical visits at the movement disorders clinic of the Service of Neurology, Montpellier Beau Soleil Clinic, France. Regulatory approvals were granted for the study (CESSRESS 22075132 Bis and CNIL~2238428). All the participants underwent video recording sessions aimed to be performed under standardized condition, camera positioning, and consistent distance between participants and the camera to minimize variability. The total duration of video recordings varied among patients due to the impact of the MDs severity on task execution and completion time. Videos were captured using smartphone cameras capable of maintaining consistent frame rates (minimum of 30 frames per second, FPS) and resolution (1920×1080 pixels). Participants performed predefined, standardized motor tasks known to elicit or exacerbate motor phenomenology, including but not limited to resting, posture-holding tasks, and voluntary movements such as finger tapping, arm flexion/extension, target reaching, and writing. The complete content of the conditions and tasks captured in the video recordings is provided as supplementary material (\textbf{Supplementary Material S1}). This acquisition setup was intentionally designed to reflect real-world clinic conditions (smartphone, routine exam).

\subsection{Clinical Assessment and Annotation Protocol}

Patients were clinically assessed by two physicians with expertise in MDs blinded to each other\textquotesingle s assessments. They independently assessed the presence or absence of eight hyperkinetic MDs: dystonia, tremor, myoclonus, chorea, athetosis, tics, ballismus, and stereotypies. Following the independent ratings, both clinicians met to review any discrepancies and establish a final consensus label (presence vs.~absence) for each hyperkinetic MD, which served as the ground truth for subsequent analyses. Ten-second temporal windows defined based on clinician annotations specifying start and end times were automatically extracted from the complete video recordings of all twenty-one enrolled subjects. Fixed 10-second windows duration has been selected as a practical compromise since long enough to provide stable statistical, spectral, and complexity features for classification, while still short enough to capture the temporal fluctuations typical of hyperkinetic MDs. Clinicians annotated video segments using a predefined rule, marking clear presence (\textquotesingle1\textquotesingle), absence (\textquotesingle0\textquotesingle), or uncertainty (\textquotesingle2\textquotesingle) for each hyperkinetic MD. Only segments marked explicitly as present or absent were included in the subsequent analysis; segments containing uncertain annotations were discarded to maintain data integrity. Data cleaning procedures were applied to remove any segments containing missing or non-numeric values. The per-phenotype frequency of uncertain and excluded windows, together with the retained class balance, is reported in \textbf{Supplementary Table S2}. \textbf{Table 1} summarizes the video data characteristics for all patients. Each hyperkinetic MD pattern was evaluated across three distinct conditions: at rest, during posture maintenance, and with actions. Healthy control subjects underwent clinical examinations to exclude any MD or other neurological or extra-neurological findings that would preclude their inclusion in the control group. The consensus label (prim\_label) served as the operational ground truth for all subsequent analyses. Inter-rater agreement at the window level prior to consensus is reported per phenotype in \textbf{Supplementary Table S3} (Cohen\textquotesingle s κ).

\subsection{Video Processing and Feature Extraction Pipeline}

Each video recording was processed individually, frame by frame, to ensure consistent and reproducible extraction of spatiotemporal features \textbf{(Fig. 1)}. Frame timestamps (30 frames per second) were derived from the video FPS to align pose time series with clinical annotations at 10-s resolution. Feature extraction was implemented in Python (version 3.10.17), leveraging established computational libraries \textsuperscript{24--32} to ensure analytical accuracy and methodological reproducibility (\textbf{Supplementary Material S4}). To quantify hyperkinetic MDs phenomenology from routine clinic videos, we relied on a keypoint-based representation of human pose, which is conceptually close to what clinicians observe at the bedside: the trajectories of recognizable anatomical landmarks over time (e.g., wrists, elbows, shoulders, hips, knees, ankles, and cranial points). Pose estimation was performed using YOLOv8 (\emph{yolov8x-pose-p6.pt}), a state-of-the-art DL architecture for 2D keypoint detection in unconstrained settings. We applied YOLOv8 2D pose estimation to obtain 17 keypoints per frame. Keypoint trajectories were stored with timestamps and used to derive displacement signals for feature computation \textbf{(Table 2)}.

\textbf{Derived Additional Time-Series Features from landmark trajectories}

Beyond raw keypoint coordinates, we derived clinically motivated time-series descriptors from each landmark trajectory to capture movement properties routinely assessed at the bedside, namely the magnitude and variability of movement, the presence of oscillatory pattern, and the degree of irregularity or complexity. For each anatomical landmark tracked in 2D with pixel coordinates \(\left( x_{t},y_{t} \right)\) at frame \(t\), we constructed a scalar displacement signal \(d_{t} = \sqrt{x_{t}^{2} + y_{t}^{2}}\) (Euclidean magnitude relative to the image origin). Trajectories were segmented into fixed-length windows aligned to expert labels, and within each window we computed a panel of descriptors designed to summarize clinically meaningful aspects of movement. First, to capture amplitude and instability, we extracted distributional summaries of \(\left\{ d_{t} \right\}\) including central tendency (mean and median), dispersion (standard deviation, variance, interquartile range, and range), and extremes (minimum and maximum). Within each 10-s window, we derived interpretable descriptors intended to match bedside constructs: postural set-point and sustained bias (distributional summaries and short-horizon rolling means), excursions and paroxysmal change (extremes and event-sensitive derivative proxies), rhythmicity (FFT peak frequency/amplitude, excluding the DC component), and irregularity/complexity beyond variance (Higuchi fractal dimension and permutation entropy). Together, these measures separate tonic deviation, oscillatory burden, abrupt jerks, and continuously evolving trajectories from the same pose-derived displacement signals. Mathematical definitions, parameter choices, and the complete feature list are provided in the \textbf{Supplementary Material S4}. For each window and each landmark displacement signal, this implementation computes 19 primary descriptors (distributional + temporal + spectral + entropy + complexity) and 3 rolling-mean descriptors, yielding 22 features per landmark per window. With \(J = 17\) landmarks, this corresponds to \(22 \times 17 = 374\) derived features per window in the multi-label pipeline. Feature names followed a transparent convention \(\left\lbrack side\rbrack\_\lbrack landmark\rbrack\_ distance\_\lbrack metric \right\rbrack\) (for example, right\_shoulder\_distance\_max or left\_knee\_distance\_perm\_entropy), facilitating direct clinical interpretation and mapping to anatomical regions.

To align explanations with clinical reasoning, we organized features into kinematic families (posture/bias, excursions, variability, rhythmicity, directionality, irregularity/complexity) and anatomical regions (cranial, upper limb, lower limb). We quantified feature importance for decision at the patient level using permutation on held-out outer test folds only, preserving the full inference pipeline (window scores, p90 pooling, and label-specific control-aware thresholds). Importance was defined as the increase in patient-level error (FP+FN)/N induced by permuting one feature and aggregated across folds using a stability criterion. We deliberately chose a generic, phenotype-agnostic kinematic feature set rather than phenotype- or task-specific clinical metrics. The targeted setting is co-occurring multi-label phenotyping across eight hyperkinetic phenomenologies and three examination conditions, for which there is no common validated panel of phenotype-specific clinical metrics. Generic descriptors based on landmark displacement preserve comparability across phenotypes and conditions, are parameter-efficient (essential in the small-cohort regime), and map onto bedside constructs such as movement burden, postural set-point, excursion, rhythmicity and complexity. We anticipate that the generic descriptors used here will be complemented in future work by phenotype-specific clinical features (spatiotemporal gait parameters, tremor-band power and centre frequency for tremor, jerk-spike statistics for myoclonus, segment-level torsion for dystonia, finger-tap kinematics for action chorea), most naturally within hierarchical pipelines combining a generic stage with phenotype-dedicated downstream modules.

\textbf{Machine Learning-Based Hyperkinetic MDs Classification}

The primary analysis was patient-level multi-label phenotyping with p90 percentile aggregation of window probabilities and label-specific control-aware thresholds, evaluated under nested cross-validation with per-label model selection within training folds, with macro-AUPRC as the primary performance metric. Patient-level sensitivity, specificity (with 95\% Wilson score confidence intervals), Hamming accuracy and macro-AUROC under nested cross-validation were prespecified supportive metrics. The window-level binary screening, patient-level aggregated binary screening, condition-specific stratification, focused four-class analysis, post-hoc per-label best (exploratory upper bound), and decision-level permutation feature-importance analyses were exploratory (\textbf{Table 3}). All classifier, preprocessing and threshold decisions were made within training folds only; outer test folds were used solely for reporting. Because multiple model families and operating points were evaluated, configurations selected as \textquotesingle best\textquotesingle{} should be regarded as upper-end estimates conditional on this cohort. Throughout the manuscript, the framework is semi-automated and annotation-guided: pose estimation and feature engineering are fully automatic, but the temporal definition of the 10-s windows and the per-window supervision labels are provided by expert clinicians, and a fully end-to-end version would additionally require automatic detection and segmentation of motor-examination tasks within long routine recordings.

\textbf{Window-Based Binary Classification of Individual Hyperkinetic MD Presence Versus Absence}

As a preliminary benchmarking step, we conducted a systematic evaluation of preprocessing and supervised learning strategies (\textbf{Supplementary Material S4}) across the eight hyperkinetic MDs phenomenologies. For each symptom, we formulated a binary window-level classification problem in which a 10-second window was labeled hyperkinetic MD-present if at least one frame within the window was annotated as symptom-positive (label = 1). To construct a conservative negative class in this screening-oriented setting, windows labeled hyperkinetic MD-absent (label = 0) were retained only for healthy controls, whereas windows labeled 0 from patients were not included. This design therefore evaluates symptom-expressing patient windows against control windows, under routine acquisition conditions. In addition, we derived a subject-level decision by aggregating out-of-fold window predictions within each subject using a majority-vote rule (≥50\% positive windows). We report subject-level correct classification rate, sensitivity, and specificity based on this aggregated decision. This aggregation reduces the influence of sporadic window-level errors when multiple windows are available per subject and provides a complementary subject-level summary alongside window-level performance. Since in this window-based step the negative class is derived exclusively from controls, results should be interpreted within this screening-versus-control framework rather than as symptom absence detection within patients.

\textbf{Patient-Level Binary Classification Using Aggregated Time Series Features.} To complement window-level pre-analyses with a subject-level summary, we implemented a patient-based binary classification framework in which each subject was represented by a single feature vector per symptom. For each recording and target symptom, we constructed a symptom time series by selecting and concatenating windows according to rater annotations. For patients, only windows in which the symptom was annotated as present were retained. The retained windows were concatenated to form a single per-subject time series per anatomical landmark, thereby focusing the representation on segments that are clinically informative for the symptom under consideration. Subjects without any retained windows for a given symptom were excluded from the corresponding symptom-specific analysis. From this concatenated signal, we extracted pose-derived time-series spanning the same feature families as the window-based pipeline yielding one aggregated feature vector per subject and symptom. Subjects were assigned a binary label reflecting group membership in this patient-versus-control setting (controls = 0; patients = 1), and models were trained to discriminate symptom-expressing patient recordings from healthy controls.

Feature rescaling was applied within each training fold. The same panel of supervised classifiers was evaluated with exhaustive grid search over predefined hyperparameter grids. For classifiers with potentially uncalibrated probability estimates, probability calibration was performed using Platt scaling (sigmoid calibration) via a calibrated classifier wrapper with internal cross-validation on the training data. Because feature vectors were derived from symptom-positive windows, this patient-level analysis characterizes a ``symptom signature'' at the subject level rather than evaluating symptom absence within patients; it therefore complements, rather than replaces, the window-level analysis.

\textbf{Condition-Specific hyperkinetic MD Detection.} In addition, to reflect clinical examination practice, we stratified recordings into three predefined conditions commonly used during examination: (i) rest, (ii) posture maintenance, and (iii) voluntary movement (e.g., finger-to-nose maneuver). For each condition and symptom, we constructed one condition-specific sample per subject by aggregating the corresponding annotated segments into a single merged time series and extracting the same pose-derived feature set. A subject--condition sample was labeled symptom-positive if the symptom was annotated as present (label = 1) in any frame within the merged series. Classifier performance was evaluated using subject-grouped cross-validation within each condition, enabling direct comparison of detection performance across motor tasks.

\subsubsection{Concurrent detection of multi-label hyperkinetic MDs}

To emulate routine clinical assessment, we evaluated the simultaneous detection of multiple hyperkinetic MDs from the 10-s video windows using an eight-label output space (dystonia, tremor, myoclonus, chorea, athetosis, ballismus, stereotypies and tics). Each window was assigned an 8-dimensional binary label vector derived from expert annotations (label = 1 if the symptom was present at any time within the window; else 0). In this multi-label setting, we explicitly controlled false alarms in asymptomatic recordings by including both, the healthy control subjects and symptom-absent windows from the patients, enabling clinically interpretable subject-level readouts under a conservative operating regime.

\textbf{Subject-level cross-validation and fold balancing.} To prevent information leakage, all windows from a given participant were assigned to the same fold. Folds were balanced using patient-level phenotype profiles (presence/absence of each of the 8 phenotypes across that participant's windows) so that individual phenotype prevalence and common co-occurrence patterns were preserved as much as possible across cross-validation splits (MultilabelStratifiedKFold; shuffle with fixed random seed).

\textbf{From window probabilities to patient-level multi-label decisions.} Each label was modelled as an independent binary task at the window level, yielding per-window probabilities for each symptom. To obtain a clinically actionable patient-level multi-label profile, window probabilities were aggregated within each patient using percentile pooling (90th percentile), producing one probability per patient and label. Decision thresholds were tuned per label on training patients only using a clinically motivated operating-point policy that minimizes a balanced error criterion under explicit constraints on false positives among control subjects (including label-specific constraints for high-impact phenotypes). Thresholds were then applied unchanged to held-out test patients. In addition, we evaluated an optional nested selection strategy in which the best-performing model configuration could be selected independently for each label within the training data (inner cross-validation), reflecting the heterogeneous learnability and risk profile of the individual phenotypes. Thus, each phenotype had its own threshold, determined from the training data only, to balance sensitivity and specificity while limiting false-positive calls in controls. Similarly, the nested strategy allowed model choice to vary by phenotype within the training data, because phenotypes differ both in detectability and in the clinical consequences of overcalling versus missing a case.

\textbf{Focused hyperkinetic MD detection: dystonia versus athetosis versus chorea versus control.} To further assess the pipeline in a clinically relevant scenario of overlapping hyperkinetic MDs observed in dyskinetic cerebral palsy (one of the most frequently acquired hyperkinetic MDs), we implemented a focused four-class classification task. Ten-second windows were annotated as dystonia, chorea, athetosis, or asymptomatic windows. Only windows with unambiguous labels were retained for analysis.

\subsection{Model training and evaluation}

Models were evaluated with subject-level cross-validation to prevent leakage. Preprocessing and model hyperparameters were selected within the training data; probability calibration was applied when needed. Full grids and implementation details are provided in Supplementary Material S4. Preprocessing consisted of feature scaling (StandardScaler, MinMaxScaler, RobustScaler, and PowerTransformer) when appropriate followed by univariate feature selection using mutual information (SelectKBest; fixed K with minor label-specific adjustments). We evaluated a panel of supervised classifiers commonly used for structured biomedical data: gradient-boosted decision trees (XGBoost, LightGBM), Random Forest, linear models (Logistic Regression, SVM with RBF kernel), instance-based learning (K-Nearest Neighbors), and a multilayer perceptron (MLP). Hyperparameters were optimized by exhaustive grid search over predefined parameter grids. Class imbalance was handled within each training fold: when the relative difference between class sizes exceeded 20\%, we applied random undersampling of the majority class to match the minority class size. For classifiers producing potentially uncalibrated probability estimates, probability calibration was applied using Platt scaling (sigmoid calibration) via a calibrated classifier wrapper with internal cross-validation on the training data. Performance was quantified at the window level using accuracy, class-specific F1-scores (reported separately for the positive and negative classes), and ROC-AUC. Specifically, for multilabel evaluation, primary performance was reported at the patient level. Discrimination was quantified using macro-averaged ROC-AUC and macro-averaged average precision (AUPRC), computed label-wise when both classes were present in the test fold and then averaged across eligible labels. For binary predictions, we reported micro- and macro-averaged F1-scores, and Hamming accuracy for the full 8-label vector. Per-label precision, recall, specificity and F1-score were also reported, and control-only false positives were tracked to verify alignment with the targeted operating regime.

\subsection{Implementation and Computation}

The pipelines described in our study were implemented using Python (version 3.10.17). All related Python scripts and results are provided on GitHub: \url{https://github.com/xaviervasques/CODY}

\section{Results}

\textbf{Window-Level Detection of Individual Hyperlinetic MDs}

Fig. 2 summarizes the best-performing window-based pipelines for each hyperkinetic MD using 10-s video segments. Throughout, these window-level binary analyses should be read as a screening contrast between symptom-expressing patient windows (≥ 1 symptom-positive frame within the 10-s window) and healthy control windows; they do not evaluate symptom absence within patient videos, which is addressed only at the patient level in the multi-label analysis. High specificity values therefore quantify how rarely a control window is misclassified as symptomatic, not the model\textquotesingle s ability to identify symptom-absent intervals inside complex patients. Across phenotypes and annotation sets, models showed consistently high subject-level specificity after within-subject aggregation (1.00 for all symptoms and raters), indicating that healthy controls were rarely labelled as symptomatic once window-level predictions were integrated at the subject level. Dystonia, chorea, tremor and tics exhibited robust window-level performance and reliable subject identification, consistent with relatively distinctive kinematic signatures: sustained abnormal posturing for dystonia, irregular purposeless excursions for chorea, rhythmic oscillatory structure for tremor, and brief stereotyped bursts for tics. In contrast, athetosis, stereotypies, myoclonus and ballismus were more challenging at the 10-s window scale, consistent with (i) slow, continuous and often mixed-pattern movements (athetosis), (ii) context-dependent and heterogeneous repetitive behaviours (stereotypies), and (iii) highly intermittent, short-duration events that can be more difficult to time-stamp consistently (myoclonus and ballismus). The lower performance for these latter phenotypes is also consistent with their lower window-level inter-rater agreement (Cohen\textquotesingle s κ = 0.116 for tics, 0.220 for ballismus, 0.259 for athetosis, 0.366 for myoclonus, and 0.000 for stereotypies; \textbf{Supplementary Table S3}) and is expected from the limitations of the 2D pose representation for fine distal and subtle facial signals.

\textbf{Symptom-Specific Performance Insights.} Dystonia demonstrated strong and consistent performance across annotation sets (F1\_1 = 0.84 vs 0.89; patient sensitivity = 0.92 vs 1.00), supporting the presence of stable pose-derived kinematic signatures that are reproducibly captured despite rater-specific labeling policies. Chorea was also robust (F1\_1 = 0.81 vs 0.96) with perfect patient-level sensitivity in both annotation sets; the larger between-annotation performance difference at the window level (ΔF1 ≈ 0.15) is consistent with subtle differences in labeling thresholds for borderline choreiform movements. Tremor performance was highly concordant across annotation sets (F1\_1 ≈ 0.82--0.83; ΔF1 ≈ 0.01) with strong patient identification (≈0.89--0.92), in keeping with a relatively rhythmic motor signature. Tics achieved excellent patient-level detection in both annotation sets (patient sensitivity = 1.00), despite greater variability at the window level for Rater 1 (F1\_1 = 0.75 ± 0.25). In contrast, athetosis yielded only moderate window-level performance (F1\_1 = 0.59 vs 0.49) with reasonable patient sensitivity (0.80 vs 0.75). Ballismus was the most challenging phenotype (F1\_1 = 0.14 vs 0.23), with marked annotation-set dependence in patient sensitivity (1.00 vs 0.50), compatible with rarity and episodic expression as well as ambiguity in defining ballistic segments. Stereotypies showed moderate performance in the single available annotation set (F1\_1 = 0.54 ± 0.26; patient sensitivity = 0.75), reflecting heterogeneous and context-dependent repetitive patterns. Finally, myoclonus was strongly annotation-set dependent (F1\_1 = 0.68 vs 0.36; patient sensitivity = 0.90 vs 0.43; ΔF1 ≈ 0.32).

\textbf{Patient-Level Classification Through Window Aggregation.} The ``percentage of subjects correctly classified'' (\textbf{Fig. 2}) quantifies subject-level discrimination between symptom-expressing patients and healthy controls obtained by aggregating out-of-fold window predictions within each subject. Each 10-s window is classified independently, and a subject is labelled ``symptom-positive'' when at least half of his windows are predicted positive (≥50\%). Clinically, this parallels a physician reviewing multiple short epochs before issuing an overall judgement based on the preponderance of evidence. Subject-level performance was consistently high across most phenotypes and annotation sets, frequently reaching 1.00, indicating that aggregation mitigates sporadic window-level errors, particularly for intermittent phenotypes, while preserving conservative behavior in controls.

\textbf{Performance Under Standardized Clinical Conditions.} To assess robustness across standardized examination contexts, we evaluated symptom-versus-control classification separately during rest, posture maintenance, and action. As summarized in \textbf{Fig. 2(B)}, dystonia and tremor remained highly discriminable across conditions for several model families, with multiple pipelines achieving near-ceiling performance. Given the smaller condition-specific subsets, these results should be interpreted as supportive evidence of context robustness.

\textbf{Multi-label detection of concurrent hyperkinetic MDs}

We next evaluated whether pose-derived kinematic features can support simultaneous, clinically interpretable detection of multiple hyperkinetic MDs, reflecting the reality that several abnormal movement patterns may co-occur within the same patient. Patient-level multi-label performance was estimated under three model-selection strategies, summarised in \textbf{Table 4}. Under the prespecified primary analysis using nested cross-validation with per-label model selection within training folds, the framework reached macro-AUPRC = 0.717 ± 0.030, macro-AUROC = 0.767 ± 0.069, Hamming accuracy = 0.764 ± 0.014, and patient--label agreement of 153/200 (76.5\%). When restricted to a single fixed pipeline selected by discrimination, the best-performing configuration (StandardScaler + MLP) achieved macro-AUPRC = 0.821 ± 0.019 and macro-AUROC = 0.830 ± 0.029 at the patient level (\textbf{Fig. 3}); when restricted to a single fixed pipeline selected by Hamming accuracy (MinMaxScaler + SVM), it achieved 0.764 ± 0.041. As an exploratory upper bound, post-hoc selection of the best-performing pipeline for each phenotype (\textbf{Table 5}) yielded patient--label agreement of 172/200 (86.0\%). The 9.5-percentage-point gap between this upper bound and the nested cross-validation estimate quantifies the model-selection contribution to apparent performance, and supports our framing of these results as exploratory rather than confirmatory. Per-phenotype sensitivity and specificity, with 95\% Wilson score confidence intervals under both the post-hoc per-label best and the nested cross-validation strategies, are reported in \textbf{Supplementary Tables S5a and S5b}.

To contextualize these aggregate metrics, we examined the best per-label models (selected to minimize patient-level errors for each phenotype) and decomposed predictions into true negatives, true positives, false positives and false negatives (\textbf{Table 5}). This exploratory upper-bound analysis comprises 200 patient--label decisions (25 subjects × 8 labels); under per-label model selection performed post-hoc, predictions agreed with clinician consensus labels in 172 decisions (86.0\%). Importantly, this 86.0\% figure is an upper-bound estimate subject to model-selection optimism, in contrast to the primary nested cross-validation estimate of 153/200 (76.5\%) reported above (\textbf{Table 4}). Performance estimates for phenotypes with very few positive participants, notably ballismus (3 positive patients), stereotypies (7) and tics (3), carry wide confidence intervals and should be regarded as exploratory signals. Under nested cross-validation, ballismus sensitivity was 0.00 (95\% Wilson CI 0.00--0.56) and tics sensitivity was 0.67 (95\% CI 0.21--0.94); under the post-hoc per-label best, the same labels reached 0.33 (95\% CI 0.06--0.79) and 1.00 (95\% CI 0.44--1.00), respectively (\textbf{Supplementary Tables S5a/S5b} Bootstrap 95\% confidence intervals for macro-AUPRC and macro-AUROC across folds are provided in \textbf{Supplementary Table S6}). The breadth of these intervals, spanning much of the {[}0,1{]} range, illustrates that with denominators of 3 to 7, even one displaced case substantially shifts the point estimate. In this cohort, tics showed apparent perfect agreement under post-hoc per-label selection (TP=3, TN=22; no discordant decisions), and dystonia demonstrated high sensitivity with only a single missed case (TP=20, FN=1) alongside minimal over-calling (FP=1). Myoclonus and chorea achieved high recall (no false negatives), with remaining discordances dominated by over-calling, an operating profile that, in practice, would be interpreted in clinical context and could be shifted by adopting stricter thresholds for phenotypes where false alarms are particularly costly. In contrast, tremor exhibited the largest discordance burden (FN=4, FP=4), consistent with borderline presentations and overlap with dystonia and non-specific movement patterns captured in short windows; athetosis similarly showed balanced under- and over-calling (FN=2, FP=3), suggesting that subtle, low-frequency fluctuations remain difficult to consolidate into an unambiguous patient-level label. For rare labels, the decomposition was particularly informative. Ballismus, observed in only three positive patients, maintained perfect specificity (FP=0) but lower sensitivity (FN=2 under post-hoc; FN=3 under nested CV), illustrating the expected behaviour of conservative detection in the low-prevalence regime. Apparent strengths or weaknesses for rare phenotypes (notably ballismus, stereotypies, and tics) cannot be regarded as established phenotype signatures from this cohort alone, and are additionally constrained by low inter-rater agreement at the window level (κ = 0.220, 0.000 and 0.116, respectively; Supplementary Table S3).

In contrast, the most clinically relative challenging phenotypes reflected familiar sources of ambiguity in routine assessment and therefore concentrated the residual discordances. Tremor exhibited the largest discordance burden (FN=4, FP=4), consistent with borderline presentations and overlap with dystonia and non-specific movement patterns captured in short windows; athetosis similarly showed balanced under- and over-calling (FN=2, FP=3), suggesting that subtle, low-frequency fluctuations remain difficult to consolidate into an unambiguous patient-level label. For rare labels, the decomposition was particularly informative. Ballismus, observed in only three positive patients, maintained perfect specificity (FP=0) but lower sensitivity (FN=2), illustrating the expected behavior of conservative detection in the low-prevalence regime: the model prioritizes avoiding over-calling a high-impact phenotype at the expense of missed cases. Overall, patient-level multi-label inference achieved high agreement.

Collectively, these focused results suggest only partial separability of dystonia, chorea, and athetosis on the basis of pose-derived kinematic summaries from short 10-s windows. Dystonia (prevalence 0.84) showed near-ceiling sensitivity with minimal false alarms, whereas chorea and athetosis remained more challenging, consistent with low prevalence (6/25 and 9/25 positive participants, respectively) and substantial phenotypic overlap in routine outpatient videos. The residual chorea--athetosis confusion is consistent with the well-known phenomenological overlap between these two phenotypes in routine clinical assessment and with the moderate-to-fair inter-rater agreement at the window level (κ = 0.600 for chorea, 0.259 for athetosis; \textbf{Supplementary Table S3}), and should not be interpreted as definitive discrimination. Larger, multicentre cohorts with adjudicated labels are needed before any clinical claim of chorea-versus-athetosis discrimination can be made. When applying the same multi-model approach to the full symptom set instead of the focused, three-label assessment, the error profile shifts in a phenotype-specific way. Chorea improves in the full panel of hyperkinetic MD labels (errors decrease from 6 → 3), suggesting that additional phenotypes and richer negative structure can offer better contrast and help isolate chorea-specific kinematic signatures. Athetosis appears broadly stable, indicating that its separability is less sensitive to whether training is restricted or expanded. In contrast, dystonia performs slightly better in the reduced panel (difference of one additional error).

\textbf{Clinically interpretable feature importance at the patient level}

To identify which kinematic descriptors most strongly influenced clinically meaningful classification decisions, we quantified patient-level permutation importance on the held-out outer test folds only, thereby preventing information leakage.

For interpretability, we summarized importance within clinically grounded kinematic families (baseline posture, sustained bias, excursions, variability, rhythmicity, and directionality) (\textbf{Fig. 4(A)}) and across coarse anatomical regions (\textbf{Fig. 4(B)}) (cranial, upper limb/proximal, lower limb/distal). Across phenotypes, importance concentrated in descriptors capturing baseline/sustained displacement and extreme excursions, whereas spectral and higher-order descriptors contributed more selectively (\textbf{Fig. 5}). Here, ``higher-order'' features mainly comprised distributional and complexity measures (e.g., skewness/kurtosis, entropy-based measures including permutation entropy, and fractal-like complexity via Higuchi fractal dimension). These features generally showed secondary, context-dependent contributions, improving discrimination in selected cases rather than acting as the primary determinants of the patient-level decision.

Anatomically, influential descriptors frequently involved cranial and proximal upper-limb landmarks, consistent with their robust visibility in standard clinical videos and their sensitive to axial/postural adjustments. Lower-limb descriptors contributed more variably: they were prominent for phenotypes with stereotyped limb patterning, and they may also become important in patients with generalized involvement (e.g., generalized dystonia), although their contribution is inherently dependent on task, camera framing, and the captured field of view. Despite shared global trends, each symptom exhibited a distinct importance profile aligned with clinical phenomenology. Dystonia was predominantly driven by baseline and sustained postural descriptors, consistent with a persistent abnormal postural set-points; excursion-type descriptors were secondary, compatible with superimposed phasic dystonic movements. The dominant topographic contribution arose from cranial/cervical and proximal upper-limb landmarks, reflecting clinically salient axial components, frequent involvement of axial and upper body segments in adulthood dystonia and compensatory upper-body adjustments. Tremor relied primarily on baseline/sustained displacement and overall movement burden rather than on a narrow spectral signature. While rhythmicity-related descriptors (FFT peak frequency/amplitude) were present, decision-level attribution suggested that tremor detection in this dataset was largely mediated through posture-linked oscillatory load and sustained variability, which is clinically plausible in a heterogeneous tremor population (including dystonic tremor), where frequency can vary across individuals, tasks, and recording conditions and where short windows can yield unstable single-peak spectral estimates. Myoclonus showed an event-like signature dominated by paroxysmal excursions (extreme minima/maxima and related excursion proxies), consistent with brief jerks producing disproportionate extremes in displacement time series; baseline descriptors contributed secondarily. Athetosis was distinguished by a comparatively stronger contribution from directionality-related descriptors (e.g., repeated reversals captured by derivative sign changes and trajectory evolution proxies), consistent with slow, writhing movement characterized by continuously evolving trajectories rather than periodic oscillation or isolated events. Stereotypies exhibited a predominantly segmental signature with comparatively greater limb (including lower-limb) involvement than most other phenotypes, consistent with repeated patterned actions that preserve a characteristic set-point while producing reproducible excursions. For rarer phenotypes, we observed plausible but less stable signatures. Chorea, ballismus, and tics tended to be driven by excursion- and variability-related descriptors, consistent with irregular, abrupt and/or large-amplitude deviations. Aggregating decision-level feature importance into clinically grounded kinematic families revealed structured overlaps between phenotypes that mirror bedside phenomenology and highlight plausible sources of diagnostic ambiguity. First, dystonia and tremor shared a strong reliance on baseline and sustained postural signals, consistent with the clinical observation that tremor often occurs on top of an abnormal postural set-point and that axial/proximal adjustments can dominate the video signature. In this context, tremor was differentiated primarily by a greater relative contribution of movement-burden descriptors (including segmental upper-limb contributors), whereas dystonia remained more strongly anchored in sustained postural bias. Second, we observed a partial overlap between myoclonus and tremor, driven by shared reliance on cranial/proximal descriptors capturing overall displacement burden. Clinically, this overlap is expected because both phenotypes can increase positional variability and intermittently produce excursions in standard video recordings; however, the relative weighting differed, with myoclonus exhibiting a more event-like profile dominated by extreme excursions (brief jerks) and tremor relying more on sustained oscillatory load (a more continuous movement pattern). Third, event-dominated phenotypes (tics, ballismus, and to some extent chorea) showed convergent importance patterns characterized by excursion- and variability-related evidence, consistent with abrupt, irregular and/or large-amplitude deviations, albeit with the above prevalence-related caveats. In contrast, athetosis emerged as the phenotype most enriched for directionality-related evidence, supporting its differentiation from both oscillatory and purely paroxysmal phenomena.

\section{Discussion}

We developed a pose-based DL framework for automatic detection of concurrent hyperkinetic MDs from outpatient videos. Previous applications explored detection, classification \textsuperscript{33}, and quantification of various specific movement disorders, including Parkinson's disease \textsuperscript{34}, gait disorders \textsuperscript{33}, ataxia \textsuperscript{35}, and tics \textsuperscript{19}. Several performance measures are typically reported, yet they are not always intuitive for clinicians \textsuperscript{36}. We report ranking-based and threshold-based patient-level metrics because they address different clinical questions: whether the model correctly prioritizes patients with likely symptom expression, and whether it makes appropriately cautious final present/absent calls when used for patient-level phenotyping. In our window-based pre-analyses, the negative class was deliberately defined conservatively as healthy control windows rather than symptom-absent windows from patients, so performance should be interpreted as discrimination between clinically labeled symptom expression and normative movement patterns, not as ``symptom absence within patients'' detection. Within this screening-oriented definition, patient-level aggregation yielded consistently high control specificity, indicating that repeated short observations rarely led to symptomatic labeling in healthy controls. This behavior is clinically desirable and suggests that pose-derived features capture stable normative movement signatures that are not misclassified as hyperkinetic activity when evidence is aggregated.

Window-level discrimination varied by phenotype, mirroring differences in phenomenology and the ease of defining symptom boundaries on routine videos. Dystonia, chorea, tremor and tics showed robust window-level reliability (positive-class F1 values in the \textasciitilde0.8--0.9 range across raters) and strong subject identification after aggregation, consistent with relatively distinctive kinematic signatures: sustained abnormal posturing for dystonia, irregular purposeless excursions for chorea, oscillatory structure for tremor, and brief, variable or stereotyped bursts for tics. In contrast, athetosis, stereotypies, myoclonus and ballismus were more challenging at the 10-second scale, consistent with (i) slow, continuous, and often mixed-pattern movements (athetosis), (ii) context-dependent repetitive behaviors (stereotypies), and (iii) highly intermittent, short-duration events that can be difficult to time-stamp consistently (myoclonus and ballismus). This pattern is reinforced by rater-dependence: tremor was highly concordant between annotation sets (ΔF1 ≈ 0.01), while myoclonus showed marked annotation-set dependence (ΔF1 ≈ 0.32), underscoring the clinical difficulty of time-stamping brief shock-like events and the sensitivity of learned signatures to labeling policy. Importantly, subject-level identification generally exceeded window-level reliability, supporting the clinical utility of aggregating repeated short observations: majority-vote consolidation reduces the influence of sporadic window-level errors and recapitulates the clinical process of forming a judgement after reviewing multiple epochs rather than relying on a single short clip.

Under standardized examination conditions (rest, posture maintenance, action), symptom-versus-control separation for dystonia and tremor remained strong across several model families, with multiple pipelines reaching near-ceiling performance across conditions.

Because clinical assessment frequently involves concurrent hyperkinetic phenotypes, we next asked whether pose-derived kinematic features can support simultaneous and clinically interpretable multi-label inference at the patient level. Discrimination was strong when probabilities were interpreted as ranked evidence rather than hard calls, indicating robust separability of symptomatic and asymptomatic subjects across cross-validation folds. Importantly, optimizing for different clinical priorities produced the expected trade-offs: maximizing overall correctness across all patient--phenotype decisions (best Hamming accuracy) yielded 0.764 ± 0.041, whereas choosing thresholds to minimize clinically meaningful misclassification led to more cautious symptom assignment, with fewer positive calls unless the evidence was stronger. This pattern is consistent with how threshold may be adapted to different clinical contexts, from initial screening to more confirmation-oriented use. This ``phenotype-specific operating landscape'' is a critical clinical message: a probabilistic backbone can be tuned to match the intended context of use rather than forcing a single operating point across all phenotypes.

To contextualize aggregate metrics, we examined best per-label models (selected to minimize patient-level errors) and decomposed predictions into true negatives, true positives, false positives and false negatives. Across 200 patient--label decisions (25 subjects × 8 labels), per-label selection agreed with clinician labels in 172/200 decisions (86.0\%). Models were more in agreement for phenotypes with distinctive and/or consistently annotated patterns: tics were perfectly identified in this cohort, and dystonia showed high sensitivity with only a single missed case and minimal overcalling. Myoclonus and chorea also showed high recall in this cohort (no false negatives), with the remaining errors mainly consisting of false positives. In practice, such false-positive calls would often be resolved by clinical assessment and may be reduced by applying stricter thresholds when a more cautious classification is preferred. Tremor carried relative disagreement, consistent with borderline presentations, task dependence, and overlap with non-specific movement patterns captured in short windows; athetosis similarly exhibited few false negatives and false positives, suggesting that subtle low-frequency fluctuations remain difficult to consolidate into an unambiguous patient-level call from brief windows alone. For rare labels, the decomposition was particularly informative: ballismus, observed in only three positive subjects, maintained perfect specificity but relative lower sensitivity, illustrating cautious detection in the low-prevalence regime, avoiding overcalling a high-impact diagnosis at the expense of missed cases.

A core methodological and clinical challenge in this domain is that no absolute ground truth exists for hyperkinetic MDs classification. As highlighted in the literature, rated tasks are vulnerable to uncertainty and inter-rater variability \textsuperscript{37}. Consequently, some apparent model ``errors'' may reflect ambiguous or borderline phenomenology rather than purely algorithmic failure; in practice, a discordant prediction can sometimes flag a segment where the reference label itself is contestable, particularly for brief, intermittent events and for phenotypes with overlapping kinematic signatures. Majority voting can provide a pragmatic pseudo-ground truth at the patient level, but it has caveats, especially when disagreement reflects true ambiguity rather than noise. Models trained on assessments from multiple assessors could potentially outperform any individual assessors in specific settings \textsuperscript{33}, but realizing this advantage will require explicit uncertainty-aware labeling (probabilistic labels, adjudication, or structured disagreement modeling) rather than excluding ambiguous cases. This issue becomes even more critical when moving from classification to quantification, where disagreement typically increases and clinical consequences of misestimation may be higher. Most current DL-based quantification work focuses on individual Parkinson's disease symptoms such as levodopa-induced dyskinesia (LID) \textsuperscript{38}, tremor, rising from a chair \textsuperscript{39}, and in the hyperkinetic domain has addressed eye tics \textsuperscript{19}, action myoclonus \textsuperscript{40} and focal dystonia \textsuperscript{41}; quantification of multiple hyperkinetic MDs remains a key next step for the field and is the focus of our ongoing work.

Across phenotypes, decision-level importance concentrated in kinematic families capturing baseline posture and sustained displacement as well as extreme excursions and variability, whereas rhythmicity and higher-order irregularity/complexity descriptors contributed more selectively. This pattern supports a clinically intuitive interpretation: the models largely rely on descriptors that correspond to what clinicians observe at the bedside, movement burden, dispersion around a postural set-point, and episodic departures, while using spectral and complexity cues to refine decisions when physiology predicts them. Importances also revealed meaningful phenotype-specific profiles: dystonia and tremor were strongly shaped by baseline and sustained postural descriptors, consistent with tonic deviation and oscillatory load on top of a postural set-point; chorea and tics were dominated by excursion-type evidence, consistent with irregular, abrupt deviations; myoclonus showed an event-like profile driven by excursion and baseline burden consistent with brief jerks that disproportionately influence extremes; and athetosis showed relatively greater contributions from directionality-related evidence, consistent with slow writhing trajectories characterized by continuous evolution and repeated reversals rather than periodic oscillation or isolated paroxysms. At the anatomical level, influential evidence frequently arose from cranial and proximal upper-limb landmarks, which are robustly visible in routine clinic videos and sensitive to axial and upper-body adjustments; lower-limb contributions were more phenotype-dependent, becoming prominent for stereotypies and for phenotypes where limb patterning is salient within the field of view.

Several considerations should be kept in mind when interpreting these results. First, the cohort is modest: 21 patients and 4 healthy controls from a single centre, with imbalance between phenotypes (n = 21, 15, 15, 9, 7, 6, 3, 3 positive participants for dystonia, tremor, myoclonus, athetosis, stereotypies, chorea, ballismus and tics, respectively). Because all 10-s windows from a given participant share a common physiology, examination context and label vector, the effective independent sample size for every analysis is the number of participants, not the number of windows. For phenotypes with single-digit numbers of positive participants (notably ballismus, stereotypies and tics), per-label estimates carry wide confidence intervals (e.g., 95\% Wilson CI for ballismus sensitivity = 0.00--0.56 under nested cross-validation; Supplementary Table S5b) and are best interpreted as exploratory signals. Although subject-level cross-validation prevents window-level leakage, within-patient windows share physiology, anatomy, examination context and label structure, and patient-level aggregation reduces noise without changing this constraint; the reported performance therefore likely overestimates what would be observed in an external, multicentre prospective setting. Second, data originate from a single centre. Third, the fixed 10-second window is a deliberate compromise that enables stable statistical and spectral descriptors but can dilute very transient events and may broaden narrowband signatures. Fourth, within each 10-s window the trajectory is summarised by interpretable, hand-crafted descriptors rather than modelled as a full temporal sequence. We chose this design deliberately: it is interpretable, parameter-efficient and well suited to the present small-cohort regime, and it captures the principal kinematic properties clinicians assess at the bedside. Because hyperkinetic movement disorders also involve onset--offset behaviour, intermittency and rhythmicity, this window-summary representation may underrepresent very brief or rapidly modulated patterns, which end-to-end temporal models could capture in future work. Fifth, inter-rater differences remain a central source of label noise. Inter-rater agreement at the window level, computed between the two independent raters prior to consensus, was substantial for dystonia (Cohen\textquotesingle s κ = 0.762) and tremor (κ = 0.724), and lower for chorea (κ = 0.600), myoclonus (κ = 0.366), athetosis (κ = 0.259) and the rare phenotypes ballismus, tics and stereotypies (κ = 0.220, 0.116 and 0.000, respectively) (Supplementary Table S3). For the rarest phenotypes, extreme prevalence imbalance depresses κ even when raw agreement is high, and the two independent raters differed most in how often they annotated stereotypies before consensus, a reminder that shared operational criteria are still maturing for these phenomenologies and the reason we adopted a consensus label as the operational ground truth. Because uncertain or contested windows were excluded before model fitting, the present task is cleaner than routine clinical use, so reported metrics are likely optimistic relative to everyday clinical ambiguity; future work adopting uncertainty-aware labelling, probabilistic targets, adjudication panels, or structured disagreement modelling, would better reflect the absence of an absolute ground truth. Sixth, our representation is 2D and is based on 17 body keypoints, omitting axial rotations, depth motion, fine distal hand movements, lip and tongue movements, and subtle facial expressions, all of which carry phenotype-specific clinical information. Phenotypes whose clinically meaningful signal depends most on these finer or out-of-plane components, axial or three-dimensional displacement (e.g., torsional dystonia), fine distal kinematics (e.g., low-amplitude action tremor, subtle distal myoclonus, choreiform finger movements), or facial activity (e.g., facial tics), would therefore benefit most from future extensions to 3D pose and to dedicated hand and face tracking; the lower nested-CV performance observed for ballismus, myoclonus and stereotypies in our cohort is consistent with this expectation. Seventh, broader and more diverse control cohorts are needed to strengthen specificity estimates and calibrate thresholds under wider real-world variability. Eighth, although subject-level fold assignment prevents window-level leakage, the breadth of the model panel and the per-label model selection introduce a residual risk of selection optimism that cannot be eliminated in a single-centre cohort of this size; configurations selected as \textquotesingle best\textquotesingle{} should be regarded as upper-end estimates conditional on this cohort. Finally, the pipeline is more accurately described as semi-automated and annotation-guided rather than fully end-to-end automatic, because the temporal definition of the 10-s windows and the per-window supervision labels are provided by expert clinicians. Establishing clinical utility was beyond the scope of this proof-of-concept study and defines our roadmap for subsequent work: external and multicentre validation, prospective evaluation, test--retest reliability assessment, and comparison against independent expert panels, followed by studies assessing whether the pipeline improves diagnosis, monitoring or trial-eligibility assessment. We therefore present these findings as a foundation for those next steps rather than as evidence of clinical readiness.

We identify five concrete methodological priorities for future work. (i) External, multicentre, prospective validation with diverse cameras, distances, lighting conditions, age ranges (including paediatric and genetically defined cohorts) and adjudicated labels, in order to quantify domain shift and recalibrate thresholds. (ii) Window-length ablation studies (e.g., 5 s, 10 s, 20 s, sliding overlap) and direct comparison of alternative temporal aggregation strategies (mean, median, percentiles, attention pooling) to identify the optimal trade-off between event capture and statistical stability. (iii) Extension to 3D pose, dedicated hand and face trackers, and multimodal inputs (synchronised audio for vocal tics, accelerometry for tremor) to address phenotypes under-served by 2D body keypoints. (iv) True temporal sequence modelling (temporal convolutional, recurrent or transformer architectures) operating on per-frame keypoint trajectories and on cross-window dependencies, to capture onset--offset behaviour, intermittency and amplitude modulation, contingent on substantially larger and multicentre cohorts. (v) Uncertainty-aware labelling and quantification (probabilistic targets, adjudication panels, structured disagreement modelling), and downstream test--retest reliability, comparison against independent expert panels, and prospective evaluation of clinical added value. In addition, symptom-specific modelling, one model per phenotype rather than a single monolithic model, appears aligned with both feature signatures and clinical reasoning, and may better accommodate phenotype-specific prevalence, ambiguity, and risk profiles, most naturally within hierarchical pipelines that combine the present generic-descriptor stage with phenotype-dedicated downstream modules incorporating task-specific clinical features (e.g., spatiotemporal gait parameters, tremor-band power and centre frequency, jerk-spike statistics for myoclonus, segment-level torsion for dystonia, finger-tap kinematics for action chorea). Workflow integration and governance will also be essential, including privacy-preserving on-device processing, real-time feedback during structured exams, and human-in-the-loop review with versioned model cards and bias audits.

\textbf{Conclusions}

In summary, this single-centre, proof-of-concept study suggests that pose-based hybrid analysis of routine clinic video may capture clinically interpretable signatures of co-occurring hyperkinetic movement disorders. The modest cohort size, the imbalanced phenotype prevalence, the single-centre design, and the absence of external or prospective validation mean that conclusions about clinical utility, generalizability, and readiness for deployment must await further validation. Translation to clinical practice will require external multicentre validation, test--retest assessment, prospective evaluation against independent expert panels, and demonstration of added value for diagnosis, monitoring or trial eligibility.

\subsection{Acknowledgements}

The study was supported by a research grant from Dystonia Medical Research Foundation Canada (Registration Number 126616598 RR0001).

\subsection{Author contributions}

Conceptualization: LC, DD, GAH, GMH, JB, XV

Methodology: LC, MB, EMM, XV

Investigation: LC, DD, GAH, JDOE, ND, MCJ, CH, TW, GMH, SHe, MB, XV

Assessment: LC, DD, GAH, JDOE, ND, MCJ, CH, TW, SHu, MD, ZS, MMUR, SHe

Data Collection: LC, DD, GAH, SHu, MD

Data Curation: LC, MB, XV

Validation: LC, XV

Writing-Review \& Editing: LC, DD, GAH, JDOE, ND, MCJ, CH, TW, GMH, SHu, MD, ZS, MMUR, SHe, MB, EMM, JB, XV

Supervision: LC, XV

\subsection{Competing interests}

D.D. received honoraria for expert opinion from TEVA. J.B. is shareholder of ONWARD Medical B.V., a company developing products for stimulation of the spinal cord, not related to this research. All other authors declare no competing interests.

\subsection{Data and materials availability}

The online version contains supplementary material available at \url{https://github.com/xaviervasques/CODY} (data and code).

\textbf{List of Supplementary Materials}

Supplementary material S1: Conditions, tasks, clinical scale.

Supplementary material S4: Methods

\textbf{Ethics statement}

This study was conducted in accordance with applicable regulations for research involving human participants and with respect for participants' privacy rights. Regulatory approvals were obtained (CESSRESS 22075132 Bis; CNIL 2238428). All participants provided written informed consent for video recording and research use. Data were handled under approved governance procedures, and analyses were performed on de-identified keypoint time series and derived kinematic features.

\textbf{\hfill\break
}

\clearpage
\section*{Tables}
\begin{landscape}
{\scriptsize
\noindent \textbf{Table 1.} \textbf{Summary of Patient Demographics, Diagnoses, and Video Data Characteristics.} Etiology was confirmed in 33\% of cases, an expected finding consistent with rates of etiologic diagnostic confirmation reported in the literature.\par\medskip
\begin{longtable}[]{@{}
  >{\raggedright\arraybackslash}p{(\columnwidth - 14\tabcolsep) * \real{0.0867}}
  >{\raggedright\arraybackslash}p{(\columnwidth - 14\tabcolsep) * \real{0.0488}}
  >{\raggedright\arraybackslash}p{(\columnwidth - 14\tabcolsep) * \real{0.0913}}
  >{\raggedright\arraybackslash}p{(\columnwidth - 14\tabcolsep) * \real{0.2321}}
  >{\raggedright\arraybackslash}p{(\columnwidth - 14\tabcolsep) * \real{0.1868}}
  >{\raggedright\arraybackslash}p{(\columnwidth - 14\tabcolsep) * \real{0.1489}}
  >{\raggedright\arraybackslash}p{(\columnwidth - 14\tabcolsep) * \real{0.1049}}
  >{\raggedright\arraybackslash}p{(\columnwidth - 14\tabcolsep) * \real{0.1004}}@{}}
\toprule\noalign{}
\begin{minipage}[b]{\linewidth}\raggedright
\textbf{Patient}
\end{minipage} & \begin{minipage}[b]{\linewidth}\raggedright
\textbf{Age (y)}
\end{minipage} & \begin{minipage}[b]{\linewidth}\raggedright
\textbf{Gender}
\end{minipage} & \begin{minipage}[b]{\linewidth}\raggedright
\textbf{Diagnostic}
\end{minipage} & \begin{minipage}[b]{\linewidth}\raggedright
\textbf{Motor phenomenology}
\end{minipage} & \begin{minipage}[b]{\linewidth}\raggedright
\textbf{Number of 10-second windows (from annotations)}
\end{minipage} & \begin{minipage}[b]{\linewidth}\raggedright
\textbf{Total video duration}
\end{minipage} & \begin{minipage}[b]{\linewidth}\raggedright
\textbf{Number of frames}
\end{minipage} \\
\midrule\noalign{}
\endhead
\bottomrule\noalign{}
\endlastfoot
1 & 51 & M & Combined dystonia & Cervical dystonia and action tremor & 97 & 16min 32s & 29787 \\
2 & 55 & M & Combined dystonia & Cervical dystonia, ataxia, myoclonus & 103 & 17min 27s & 31499 \\
3 & 27 & F & Hypoxic cerebral palsy & Generalized dystonia, spasticity, athetosis & 131 & 22min 00s & 39636 \\
4 & 36 & M & \emph{DIP2B} gene related mixed movement disorder & Generalized dystonia, chorea, ataxia and athetosis & 150 & 25min 35s & 46095 \\
5 & 34 & M & \emph{DIP2B} gene related mixed movement disorder & Generalized dystonia, chorea, with ataxia & 121 & 20min 19s & 36617 \\
6 & 69 & F & Combined dystonia & Progressive cranial dystonia, mild ataxia, action tremor & 107 & 18min 09s & 32695 \\
7 & 25 & M & Hyperkinetic movement disorders & Myoclonus, tics, tremor, dytonia & 109 & 18min 31s & 33355 \\
8 & 60 & F & \emph{LRRK2} gene related Parkinson's disease & Parkinsonism and dystonia & 97 & 16min 16s & 29323 \\
9 & 73 & F & Dystonia & Generalized dystonia, mild ataxia & 114 & 19min 02s & 34288 \\
10 & 71 & F & Chorea-dystonia & Generalized chorea, focal dystonia & 127 & 21min 20s & 38396 \\
11 & 20 & F & \emph{ADCY5} gene related mixed movement disorder & Generalized dystonia, chorea, myoclonus & 95 & 15min 51s & 28556 \\
12 & 75 & F & Chorea and ataxia & Generalized dystonia, chorea, ataxia & 115 & 19min 16s & 34702 \\
13 & 62 & F & Dystonia & Generalized dystonia & 104 & 17min 24s & 31337 \\
14 & 22 & M & Tremor prominent movement disorder & Action tremor, dystonia & 108 & 18min 01s & 32456 \\
15 & 63 & F & Chorea-dystonia & Dystonia, chorea & 115 & 19min 18s & 34736 \\
16 & 69 & F & Combined dystonia & Generalized dystonia, mild parkinsonism & 105 & 17min 36s & 31699 \\
17 & 70 & F & Dystonia & Generalized dystonia & 114 & 19min 07s & 34433 \\
18 & 37 & M & \emph{TTPA} gene related ataxia with vitamin E deficiency (AVED) & Ataxia, cervical dystonia & 119 & 19min 58s & 35965 \\
19 & 26 & M & Neurodevelopmental disorder and tardive dystonia & Dystonia, akathisia & 105 & 17 min 22 s & 31275 \\
20 & 17 & F & Myoclonus dystonia & Dystonia, myoclonus & 99 & 16 min 27 s & 29618 \\
21 & 23 & M & \emph{TIMM8} gene related combined movement disorder & Generalized dystonia, tremor, spasticity, tics & 102 & 16 min 54 s & 30428 \\
\end{longtable}
}
\end{landscape}

\bigskip

{\small
\noindent \textbf{Table 2. Overview of Extracted Features from Video Pose Estimation and Time Series Analysis.} This table summarizes the different types of features extracted from the pose estimation and time series processing pipeline applied to each video recording. For each feature type, the table reports the number of features computed, their specific names or examples, and a brief description of their methodological role in the dataset. This multi-level feature extraction approach enables a comprehensive and physiologically meaningful characterization of movement dynamics for subsequent machine learning analysis.\par\medskip
\begin{longtable}[]{@{}
  >{\raggedright\arraybackslash}p{(\columnwidth - 8\tabcolsep) * \real{0.1867}}
  >{\raggedright\arraybackslash}p{(\columnwidth - 8\tabcolsep) * \real{0.0971}}
  >{\raggedright\arraybackslash}p{(\columnwidth - 8\tabcolsep) * \real{0.2030}}
  >{\raggedright\arraybackslash}p{(\columnwidth - 8\tabcolsep) * \real{0.2096}}
  >{\raggedright\arraybackslash}p{(\columnwidth - 8\tabcolsep) * \real{0.3036}}@{}}
\toprule\noalign{}
\begin{minipage}[b]{\linewidth}\raggedright
\textbf{Feature type}
\end{minipage} & \begin{minipage}[b]{\linewidth}\raggedright
\textbf{\#}
\end{minipage} & \begin{minipage}[b]{\linewidth}\raggedright
\textbf{Examples / names}
\end{minipage} & \begin{minipage}[b]{\linewidth}\raggedright
\textbf{Methodological role}
\end{minipage} & \begin{minipage}[b]{\linewidth}\raggedright
\textbf{Clinical interpretation (bedside meaning)}
\end{minipage} \\
\midrule\noalign{}
\endhead
\bottomrule\noalign{}
\endlastfoot
Anatomical landmarks (2D keypoints) & 17 & Nose, eyes, ears, L/R shoulder, elbow, wrist, hip, knee, ankle & Frame-wise detection of standardized body landmarks & Topography: where movement occurs (distal vs proximal; cranial vs appendicular involvement) \\
Landmark displacement signals (image-plane) & 17 & d\_t = sqrt(x\_t\^{}2 + y\_t\^{}2) per landmark & Converts 2D trajectories into robust scalar time series per segment & Movement amplitude proxy: how much each body part visibly moves over time \\
Statistical descriptors & 12 & mean, std, median, min/max, range, IQR, variance, skewness, kurtosis, energy, zero-crossings & Summarize distribution and variability of each displacement signal & Severity/variability: sustained large amplitudes vs fluctuating/bursty motion; frequent reversals can suggest oscillatory/intermittent patterns \\
Temporal / dynamic features & 3 (+ rolling means) & slope, mean absolute acceleration, rolling means (w = 3, 5, 7) & Capture trends, abrupt changes, and short-term variability & Abruptness/intermittence: jerk or burst -like events (myoclonus, tics), progressive change during tasks (task exacerbation/fatigability) \\
Spectral features & 2 & dominant frequency, peak amplitude (FFT) & Quantify periodicity/oscillation of displacement signals & Rhythmicity: presence of a dominant oscillatory component consistent with tremor-like patterns or repeated stereotyped movements \\
Non-linear / complexity features & 2 & Higuchi fractal dimension, permutation entropy & Quantify irregularity/complexity beyond simple variance & Irregular/chaotic motion: higher complexity often aligns with non-rhythmic hyperkinesia (e.g., chorea-like components) vs more stereotyped patterns \\
Metadata & 4 & timestamp, frame index, video ID, condition (Rest/Posture/Action) & Temporal alignment and stratified analyses & Context of observation: rest vs posture vs action for clinically meaningful stratification \\
\end{longtable}
}

\bigskip

\begin{landscape}
{\footnotesize
\noindent \textbf{Table 3. Study-Design Overview.}

\emph{For each analysis, the unit of evaluation, the definition of the negative class, the aggregation rule used to combine window-level outputs (when applicable), and the output and primary metric are summarised. The PRIMARY analysis is highlighted; all other analyses are explicitly labelled as exploratory.}\par\medskip
\begin{longtable}[]{@{}
  >{\raggedright\arraybackslash}p{(\columnwidth - 8\tabcolsep) * \real{0.1897}}
  >{\raggedright\arraybackslash}p{(\columnwidth - 8\tabcolsep) * \real{0.1466}}
  >{\raggedright\arraybackslash}p{(\columnwidth - 8\tabcolsep) * \real{0.1983}}
  >{\raggedright\arraybackslash}p{(\columnwidth - 8\tabcolsep) * \real{0.2328}}
  >{\raggedright\arraybackslash}p{(\columnwidth - 8\tabcolsep) * \real{0.2328}}@{}}
\toprule\noalign{}
\begin{minipage}[b]{\linewidth}\raggedright
\textbf{Analysis}
\end{minipage} & \begin{minipage}[b]{\linewidth}\raggedright
\textbf{Unit of evaluation}
\end{minipage} & \begin{minipage}[b]{\linewidth}\raggedright
\textbf{Negative-class definition}
\end{minipage} & \begin{minipage}[b]{\linewidth}\raggedright
\textbf{Aggregation rule}
\end{minipage} & \begin{minipage}[b]{\linewidth}\raggedright
\textbf{Output / Primary metric}
\end{minipage} \\
\midrule\noalign{}
\endhead
\bottomrule\noalign{}
\endlastfoot
Window-level binary screening (exploratory; \textbf{Figure 2}) & 10-s window & Control windows only (patient symptom-absent windows excluded) & None at window level; majority vote ≥50\% for patient summary & Window F1\_1, ROC-AUC; patient summary: \% correct, sensitivity, specificity \\
Patient-level binary, symptom-positive windows (exploratory) & Participant & Healthy controls & Concatenation of symptom-positive windows → 1 feature vector per participant & Patient accuracy, F1, ROC-AUC \\
Condition-specific stratification (rest / posture / action; exploratory) & Participant within condition & Healthy controls within the same condition & Concatenation per condition & F1 within condition \\
PRIMARY: Patient-level multi-label phenotyping (nested cross-validation) & Participant × 8 labels & Healthy controls + patient symptom-absent windows & p90 percentile of window probabilities; control-aware per-label thresholds tuned on training participants; nested per-label model selection within training folds & Macro-AUPRC (primary); supportive: macro-AUROC, Hamming accuracy, per-label sensitivity/specificity with 95\% Wilson CI \\
Exploratory upper-bound: post-hoc per-label best & Participant × 8 labels & Same as primary & Same as primary, but the best model is selected post-hoc per phenotype to minimise errors & Patient--label agreement (172/200); per-label confusion (\textbf{Supplementary Table S5a}) \\
Focused 4-class (dystonia / chorea / athetosis / controls; exploratory) & 10-s window with unambiguous label & Asymptomatic control windows & Per-label window probability & F1, AUROC, AUPRC per class \\
Permutation feature importance (exploratory) & Participant × label & --- & Permute feature → recompute patient-level error using full inference pipeline & Δerror share by kinematic family / anatomical region (\textbf{Figs. 4--5}) \\
\end{longtable}
}
\end{landscape}

\bigskip

{\small
\noindent \textbf{Table 4. Patient-level multi-label performance across model-selection strategies.}

\emph{Comparison of patient-level multi-label classification performance under four model-selection strategies (3-fold subject-level cross-validation, n = 25 participants, 8 phenotypes). The primary analysis (highlighted) is the prespecified nested cross-validation with per-label model selection performed inside the training folds. The other strategies are exploratory and are reported here only to make explicit the contribution of model-selection optimism to apparent performance: the 9.5-percentage-point gap between the post-hoc per-label upper bound (172/200 = 86.0\%) and the nested cross-validation estimate (153/200 = 76.5\%) quantifies this optimism. Values are mean ± standard deviation across folds; per-phenotype sensitivity and specificity with 95\% Wilson confidence intervals are reported in Supplementary Tables S5a (post-hoc) and S5b (nested CV).}\par\medskip
\begin{longtable}[]{@{}
  >{\raggedright\arraybackslash}p{(\columnwidth - 8\tabcolsep) * \real{0.4035}}
  >{\raggedright\arraybackslash}p{(\columnwidth - 8\tabcolsep) * \real{0.1491}}
  >{\raggedright\arraybackslash}p{(\columnwidth - 8\tabcolsep) * \real{0.1491}}
  >{\raggedright\arraybackslash}p{(\columnwidth - 8\tabcolsep) * \real{0.1491}}
  >{\raggedright\arraybackslash}p{(\columnwidth - 8\tabcolsep) * \real{0.1491}}@{}}
\toprule\noalign{}
\begin{minipage}[b]{\linewidth}\raggedright
\textbf{Strategy}
\end{minipage} & \begin{minipage}[b]{\linewidth}\raggedright
\textbf{macro-AUPRC}
\end{minipage} & \begin{minipage}[b]{\linewidth}\raggedright
\textbf{macro-AUROC}
\end{minipage} & \begin{minipage}[b]{\linewidth}\raggedright
\textbf{Hamming accuracy}
\end{minipage} & \begin{minipage}[b]{\linewidth}\raggedright
\textbf{Patient--label agreement}
\end{minipage} \\
\midrule\noalign{}
\endhead
\bottomrule\noalign{}
\endlastfoot
Best single pipeline by macro-AUPRC (\allowbreak{}StandardScaler +\allowbreak{} MLP)\allowbreak{}~: exploratory & 0.821 ± 0.019 & 0.830 ± 0.029 & 0.731 ± 0.036 & 146/200 (\allowbreak{}73.0\%)\allowbreak{} \\
Best single pipeline by Hamming accuracy (\allowbreak{}MinMaxScaler +\allowbreak{} SVM)\allowbreak{}~: exploratory & 0.769 ± 0.049 & 0.749 ± 0.094 & 0.764 ± 0.041 & 153/200 (\allowbreak{}76.5\%)\allowbreak{} \\
Nested cross-validation,\allowbreak{} per-label best within training folds~: PRIMARY & 0.717 ± 0.030 & 0.767 ± 0.069 & 0.764 ± 0.014 & 153/200 (\allowbreak{}76.5\%)\allowbreak{} \\
Post-hoc per-label best (\allowbreak{}upper bound)\allowbreak{} --- exploratory & --- & --- & --- & 172/200 (\allowbreak{}86.0\%)\allowbreak{} \\
\end{longtable}
}

\bigskip

\begin{landscape}
{\scriptsize
\noindent \textbf{Table 5.} Best per-label patient-level classifiers and confusion-matrix breakdown (multi-label, p90 aggregation).~For each HMD phenotype, we report the pipeline with the fewest patient-level errors. We provide P/N counts and prevalence, patient-level TP/FN/FP/TN, FNR and FPR, total errors (FP+FN), errors per subject, and recall/specificity.\par\medskip
\begin{longtable}[]{@{}
  >{\raggedright\arraybackslash}p{(\columnwidth - 28\tabcolsep) * \real{0.2382}}
  >{\raggedright\arraybackslash}p{(\columnwidth - 28\tabcolsep) * \real{0.0785}}
  >{\raggedright\arraybackslash}p{(\columnwidth - 28\tabcolsep) * \real{0.0346}}
  >{\raggedright\arraybackslash}p{(\columnwidth - 28\tabcolsep) * \real{0.0377}}
  >{\raggedright\arraybackslash}p{(\columnwidth - 28\tabcolsep) * \real{0.0880}}
  >{\raggedright\arraybackslash}p{(\columnwidth - 28\tabcolsep) * \real{0.0385}}
  >{\raggedright\arraybackslash}p{(\columnwidth - 28\tabcolsep) * \real{0.0393}}
  >{\raggedright\arraybackslash}p{(\columnwidth - 28\tabcolsep) * \real{0.0377}}
  >{\raggedright\arraybackslash}p{(\columnwidth - 28\tabcolsep) * \real{0.0401}}
  >{\raggedright\arraybackslash}p{(\columnwidth - 28\tabcolsep) * \real{0.0498}}
  >{\raggedright\arraybackslash}p{(\columnwidth - 28\tabcolsep) * \real{0.0483}}
  >{\raggedright\arraybackslash}p{(\columnwidth - 28\tabcolsep) * \real{0.0621}}
  >{\raggedright\arraybackslash}p{(\columnwidth - 28\tabcolsep) * \real{0.0636}}
  >{\raggedright\arraybackslash}p{(\columnwidth - 28\tabcolsep) * \real{0.0588}}
  >{\raggedright\arraybackslash}p{(\columnwidth - 28\tabcolsep) * \real{0.0848}}@{}}
\toprule\noalign{}
\begin{minipage}[b]{\linewidth}\raggedright
\textbf{Model}
\end{minipage} & \begin{minipage}[b]{\linewidth}\raggedright
\textbf{Label}
\end{minipage} & \begin{minipage}[b]{\linewidth}\raggedright
\textbf{P}
\end{minipage} & \begin{minipage}[b]{\linewidth}\raggedright
\textbf{N}
\end{minipage} & \begin{minipage}[b]{\linewidth}\raggedright
\textbf{Prevalence}
\end{minipage} & \begin{minipage}[b]{\linewidth}\raggedright
\textbf{TP}
\end{minipage} & \begin{minipage}[b]{\linewidth}\raggedright
\textbf{FN}
\end{minipage} & \begin{minipage}[b]{\linewidth}\raggedright
\textbf{FP}
\end{minipage} & \begin{minipage}[b]{\linewidth}\raggedright
\textbf{TN}
\end{minipage} & \begin{minipage}[b]{\linewidth}\raggedright
\textbf{FNR}
\end{minipage} & \begin{minipage}[b]{\linewidth}\raggedright
\textbf{FPR}
\end{minipage} & \begin{minipage}[b]{\linewidth}\raggedright
\textbf{Errors}
\end{minipage} & \begin{minipage}[b]{\linewidth}\raggedright
\textbf{Errors per patient}
\end{minipage} & \begin{minipage}[b]{\linewidth}\raggedright
\textbf{Recall}
\end{minipage} & \begin{minipage}[b]{\linewidth}\raggedright
\textbf{Specificity}
\end{minipage} \\
\midrule\noalign{}
\endhead
\bottomrule\noalign{}
\endlastfoot
MinMaxScaler+\allowbreak{}SVM(\allowbreak{}C=3.0,\allowbreak{}gamma=0.03)\allowbreak{} & Dystonia & 21 & 4 & 0.84 & 20 & 1 & 1 & 3 & 0.05 & 0.25 & 2 & 0.08 & 0.95 & 0.75 \\
StandardScaler\_+\allowbreak{}SVM(\allowbreak{}C=3.0,\allowbreak{}gamma=scale)\allowbreak{} & Tremor & 15 & 10 & 0.6 & 11 & 4 & 4 & 6 & 0.27 & 0.40 & 8 & 0.32 & 0.73 & 0.60 \\
MinMaxScaler+\allowbreak{}KNN(\allowbreak{}k=9,\allowbreak{}w=uniform,\allowbreak{}p=2)\allowbreak{} & Myoclonus & 15 & 10 & 0.6 & 15 & 0 & 4 & 6 & 0 & 0.40 & 4 & 0.16 & 1 & 0.60 \\
MinMaxScaler+\allowbreak{}KNN(\allowbreak{}k=9,\allowbreak{}w=uniform,\allowbreak{}p=2)\allowbreak{} & Chorea & 6 & 19 & 0.24 & 6 & 0 & 3 & 16 & 0 & 0.16 & 3 & 0.12 & 1 & 0.84 \\
MinMaxScaler+\allowbreak{}KNN(\allowbreak{}k=9,\allowbreak{}w=uniform,\allowbreak{}p=2)\allowbreak{} & Athetosis & 9 & 16 & 0.36 & 7 & 2 & 3 & 13 & 0.22 & 0.19 & 5 & 0.20 & 0.78 & 0.81 \\
StandardScaler+\allowbreak{}SVM(\allowbreak{}C=3.0,\allowbreak{}gamma=scale)\allowbreak{} & Ballismus & 3 & 22 & 0.12 & 1 & 2 & 0 & 22 & 0.67 & 0 & 2 & 0.08 & 0.33 & 1 \\
StandardScaler+\allowbreak{}SVM(\allowbreak{}C=10.0,\allowbreak{}gamma=0.02)\allowbreak{} & Stereotypies & 7 & 18 & 0.28 & 6 & 1 & 3 & 15 & 0.14 & 0.17 & 4 & 0.16 & 0.86 & 0.83 \\
MinMaxScaler+\allowbreak{}SVM(\allowbreak{}C=1.0,\allowbreak{}gamma=scale)\allowbreak{} & Tics & 3 & 22 & 0.12 & 3 & 0 & 0 & 22 & 0 & 0 & 0 & 0 & 1 & 1.00 \\
\end{longtable}
}
\end{landscape}

\clearpage
\section*{Figures}
\begin{figure}[p]
\centering
\includegraphics[width=\textwidth,height=0.85\textheight,keepaspectratio]{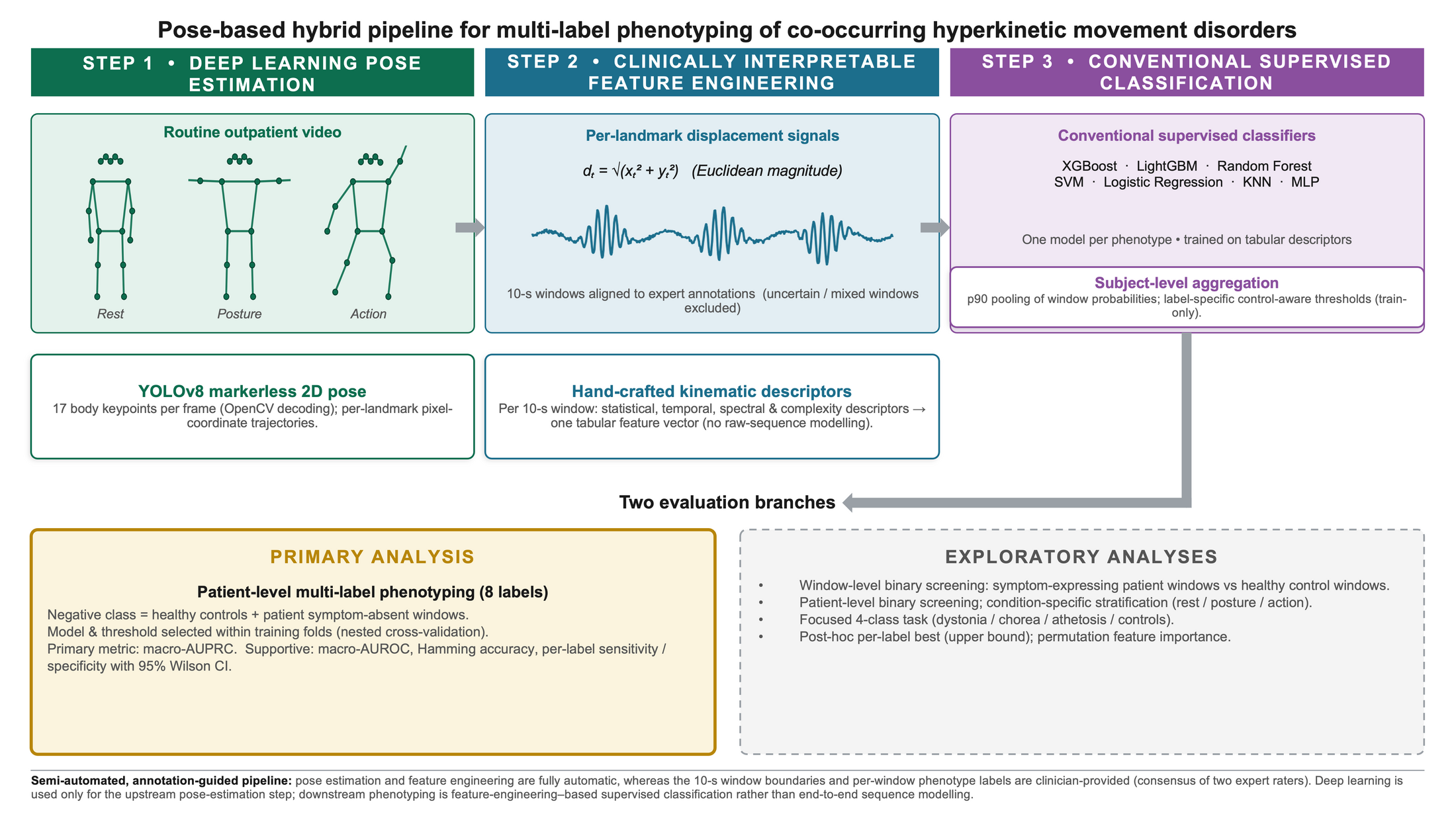}
\caption*{Fig. 1. Video-to-classification pipeline for semi-automated, annotation-guided HMD phenotyping.~Clinical videos are decoded with OpenCV and analyzed with YOLOv8 (the deep learning pose-estimation step) to estimate 17 2D keypoints per frame. Keypoints are converted to pixel coordinates, transformed into displacement signals, and segmented into consecutive 10-s windows aligned to expert window-level annotations. For each window, statistical, temporal, spectral, and complexity descriptors are computed to form a feature vector (the feature-engineering step), which is then classified by conventional supervised models (the classification step); window boundaries and labels are clinician-provided, hence "annotation-guided". Uncertain segments are excluded. Two datasets are derived: (i) binary screening (symptom-expressing patient windows vs control windows) and (ii) an 8-label multi-label set with per-window symptom vectors. In the multi-label branch, one model is trained per label; window probabilities are aggregated to the subject level (e.g., p90) and label-specific thresholds are tuned on training subjects under control-aware constraints. Performance is reported with ROC-AUC/AUPRC and multi-label metrics (Hamming, Jaccard). Example images use a consented healthy control.}
\end{figure}

\begin{figure}[p]
\centering
\includegraphics[width=\textwidth,height=0.85\textheight,keepaspectratio]{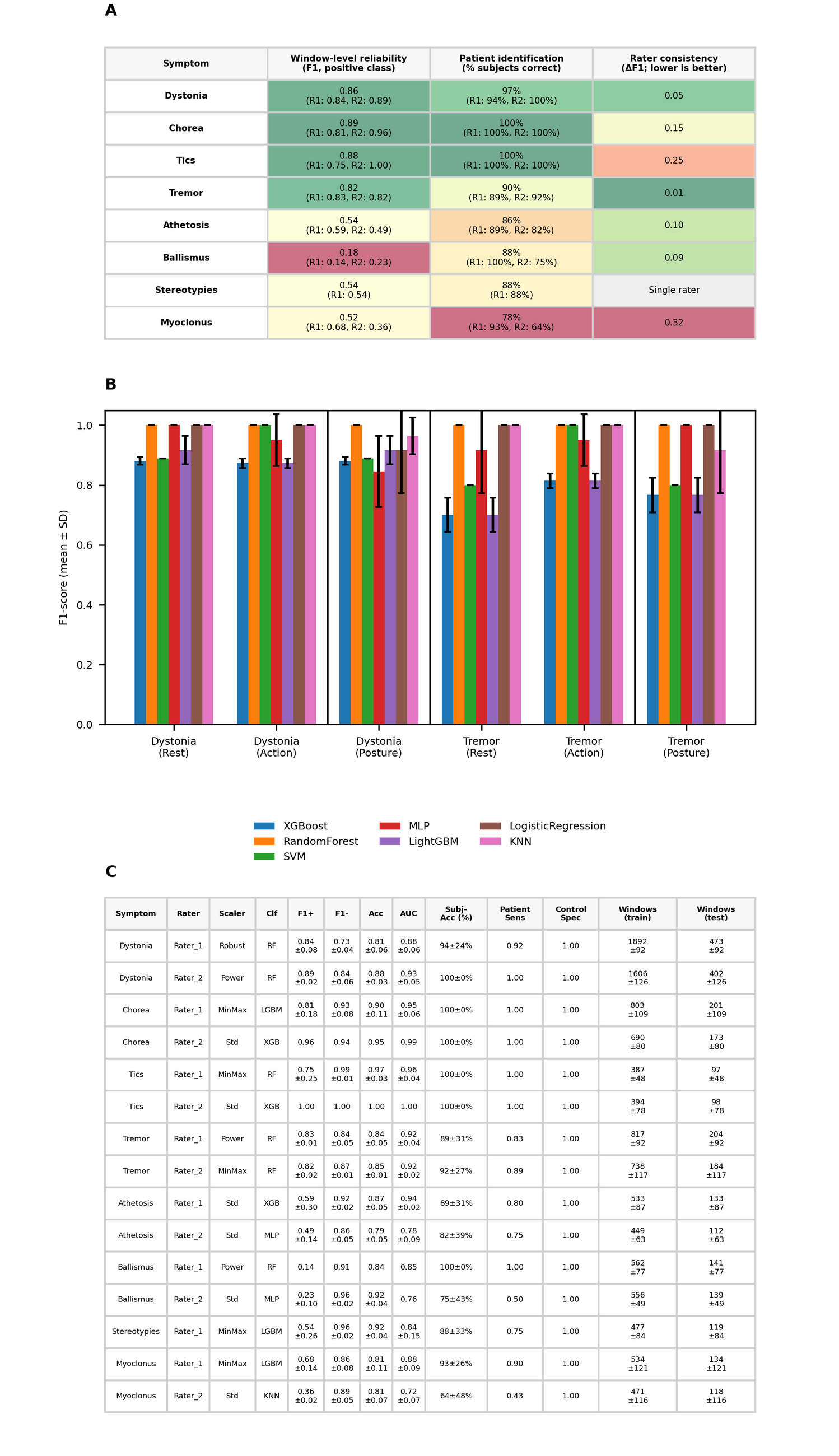}
\caption*{\textbf{Fig. 2.} \textbf{Window-based symptom classification performance across rest, posture, and action.~(A)}~Summary of window-based binary symptom detection per phenotype, reporting mean positive-class F1 (with rater-specific values), patient-level identification rate (from aggregated window predictions), and inter-rater consistency (ΔF1; lower = higher agreement). Cell colors provide a qualitative cue (green high; orange intermediate; red low; for ΔF1, green low disagreement).~\textbf{(B)} Condition-specific performance on 10-s segments (Rest, Posture, Action): mean positive-class F1 ± SD across folds for one representative model per classifier family, enabling comparison across model families and motor contexts. \textbf{(C)} For each phenotype and rater (two independent annotation sets), we report the best scaler-classifier pipeline. Performance was evaluated with 5-fold StratifiedGroupKFold cross-validation with subject-level splitting. Window-level metrics (F1\_1, F1\_0, accuracy, ROC-AUC) are mean ± SD across folds. Subject-level performance is obtained by majority voting on out-of-fold window predictions (≥50\% positive) and reports percent correctly classified subjects, sensitivity in symptom-expressing patients, and specificity in controls. Mean ± SD train/test window counts are shown. The negative class includes controls only.}
\end{figure}

\begin{figure}[p]
\centering
\includegraphics[width=\textwidth,height=0.85\textheight,keepaspectratio]{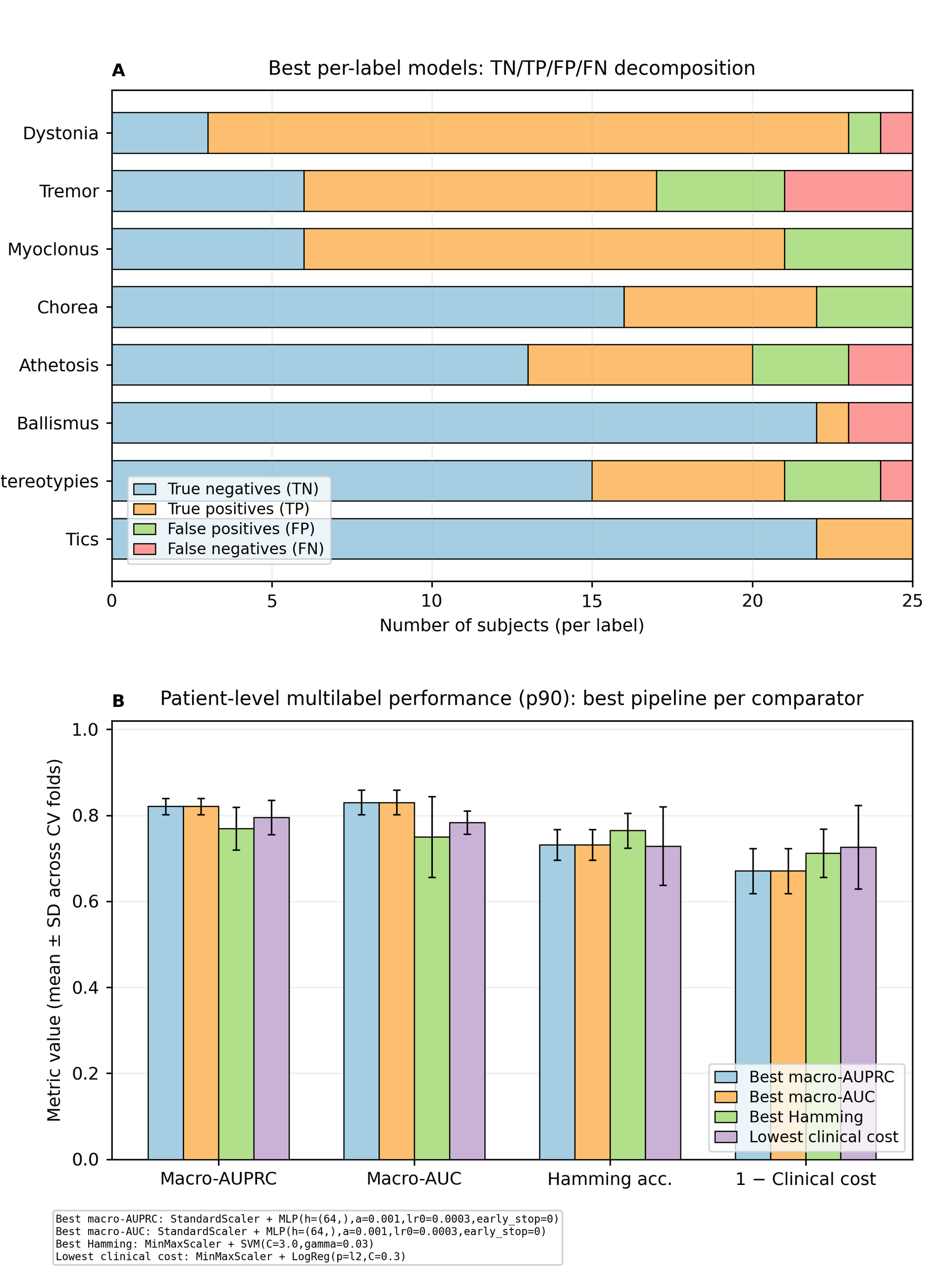}
\caption*{\textbf{Fig. 3.} \textbf{Patient-level multi-label detection of HMDs using p90 aggregation of window probabilities.}~\textbf{(A)}~Subject-level TN/TP/FP/FN counts for the best per-label pipelines (TN+TP+FP+FN per label), highlighting error types across phenotypes in the context of label prevalence.~\textbf{(B)}~Best overall pipelines under different selection criteria (macro-AUPRC, macro-AUC, Hamming accuracy, 1−clinical cost), reported as mean ± s.d. across folds, illustrating trade-offs between discrimination and error-sensitive objectives under the same p90 aggregation and per-label thresholding.}
\end{figure}

\begin{figure}[p]
\centering
\includegraphics[width=\textwidth,height=0.85\textheight,keepaspectratio]{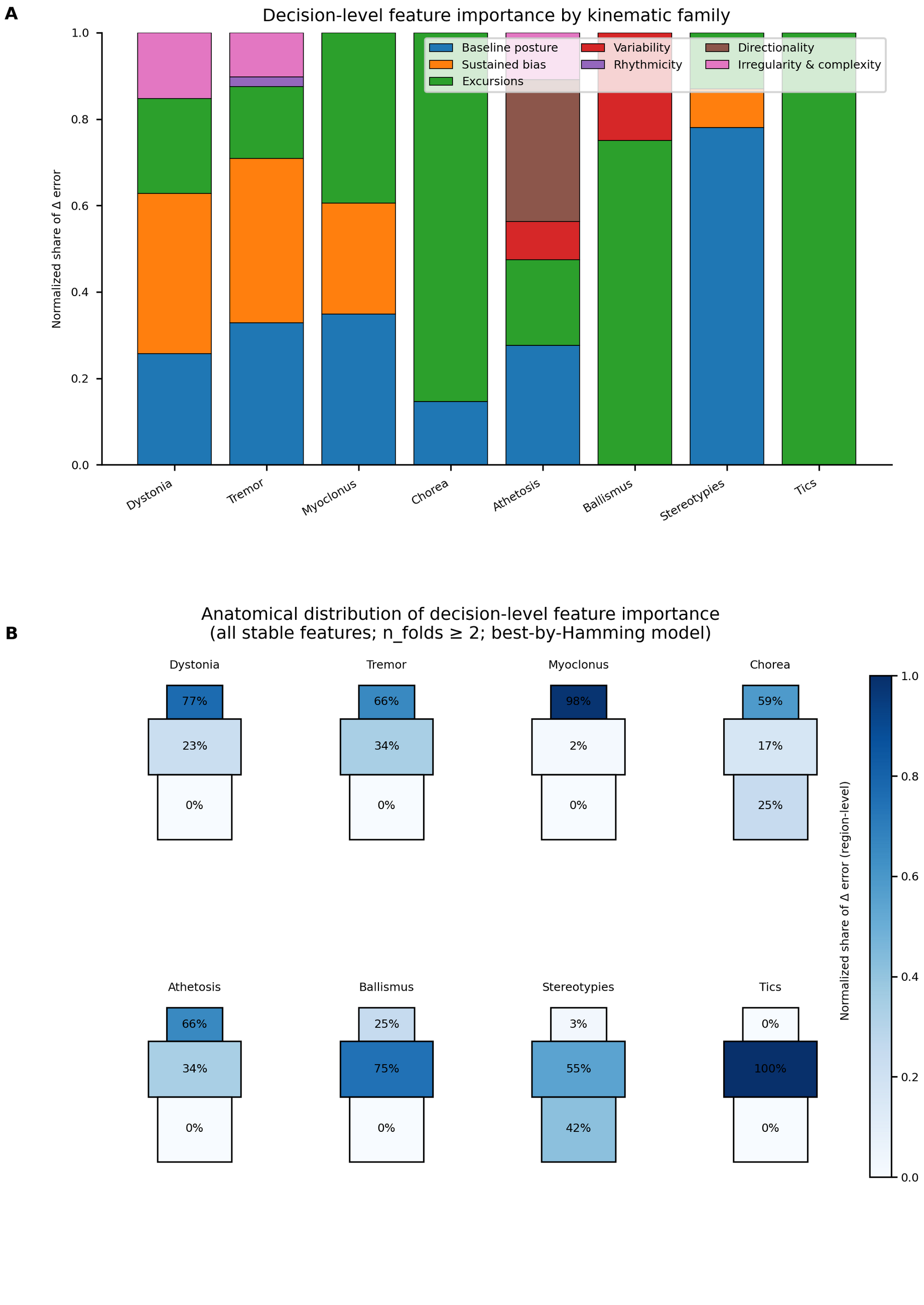}
\caption*{\textbf{Fig. 4. Patient-level decision-level feature importance.}~\textbf{(A)}~Permutation importance by kinematic family for the best-by-Hamming model (MinMaxScaler+SVM; C=3.0, γ=0.03), computed on held-out outer folds (n\_folds≥2). Importance is the increase in patient-level error (FP+FN)/N after permuting each feature, preserving the full pipeline (window scores → p90 aggregation → label-specific thresholds). Bars show the normalized Δerror share per kinematic family within each symptom. Importance is dominated by postural set-point and excursion features, with more selective contributions from rhythmicity and irregularity/complexity.~\textbf{(B)}~The same attributions aggregated by landmark region (head/face, upper limb, lower limb) as normalized Δerror share per symptom, highlighting predominant head/face and proximal upper-limb evidence, with phenotype-dependent lower-limb contributions.}
\end{figure}

\begin{figure}[p]
\centering
\includegraphics[width=\textwidth,height=0.85\textheight,keepaspectratio]{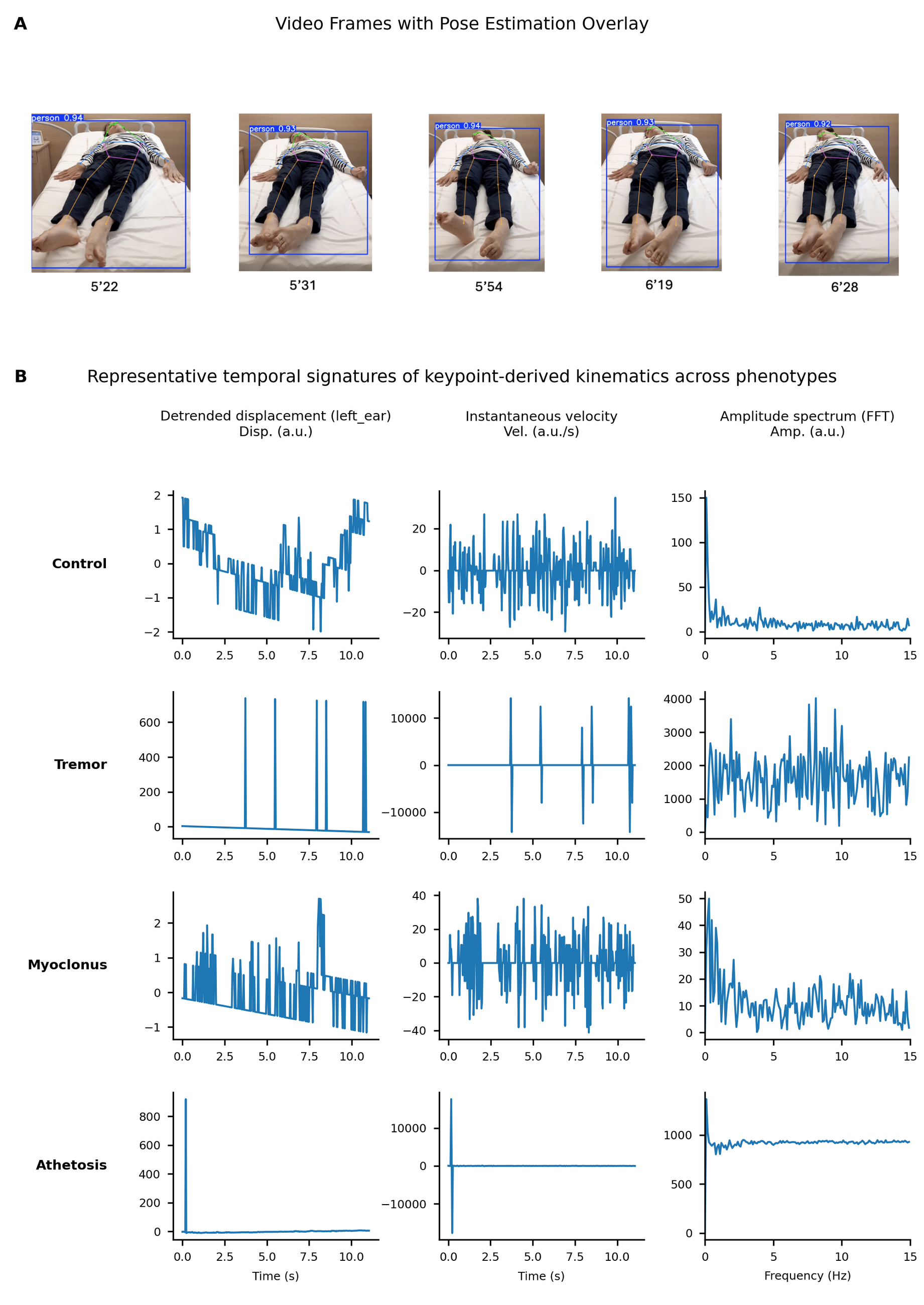}
\caption*{\textbf{Fig. 5. Pose estimation and kinematic signatures across phenotypes.}~\textbf{(A)} Example video frames with YOLOv8 pose-estimation overlay. \textbf{(B)} Representative keypoint-derived kinematic signatures extracted from the selected landmark (here, the left ear) for one control and three phenotypes. Three complementary signals are shown for each example: detrended displacement (Disp., arbitrary units {[}a.u.{]}), defined as the landmark displacement relative to its median position after linear detrending; instantaneous velocity (Vel., a.u./s), computed as the first derivative of the detrended displacement; and the amplitude spectrum (Amp., a.u.) obtained by fast Fourier transform (FFT). Controls show low-amplitude fluctuations without a dominant spectral peak. Tremor shows sustained oscillatory burden with repeated velocity alternations and enhanced spectral power. Myoclonus shows abrupt excursions and sharp velocity spikes, consistent with brief, jerky movements. Athetosis shows slower evolving fluctuations with frequent directional changes, consistent with trajectory-evolution features.}
\end{figure}

\end{document}